\documentclass{article}
\usepackage{ijcai25}

\usepackage{times}
\usepackage{soul}
\usepackage{xcolor}
\usepackage{url}
\usepackage[hidelinks]{hyperref}
\usepackage[utf8]{inputenc}
\usepackage[small]{caption}
\usepackage{graphicx}
\usepackage{amsmath}
\usepackage{amsthm}
\usepackage{booktabs}
\usepackage{algorithm}
\usepackage{algorithmic}
\usepackage[switch]{lineno}

\usepackage{multirow}
\usepackage{amssymb}
\usepackage{color}
\usepackage{enumerate}
\usepackage{xspace}
\usepackage{wrapfig}
\usepackage{natbib} 

\newcommand{\method}{MagicTailor\xspace}
\newcommand{\cred}[1]{{\textcolor[RGB]{197, 0, 19}{#1}}}
\newcommand{\cblue}[1]{{\textcolor[RGB]{3, 96, 128}{#1}}}
\newcommand{\ie}{\textit{i.e.}}
\newcommand{\eg}{\textit{e.g.}}
\definecolor{cvprblue}{rgb}{0.21,0.49,0.74}

\newcommand\blfootnote[1]{%
  \begingroup
  \renewcommand\thefootnote{}\footnote{#1}%
  \addtocounter{footnote}{-1}%
  \endgroup
}

\title{\method: Component-Controllable Personalization \\ in Text-to-Image Diffusion Models}

\author{
Donghao Zhou$^{1*}$ \and
Jiancheng Huang$^{2*}$ \and
Jinbin Bai$^{3}$ \and
Jiaze Wang$^{1}$ \and
Hao Chen$^{1}$ \and \\
Guangyong Chen$^{4}$ \and
Xiaowei Hu$^{5\dagger}$ \and
Pheng-Ann Heng$^{1}$ \vspace{1.5mm} 
\affiliations
$^{1}$CUHK \qquad
$^{2}$SIAT, CAS \qquad
$^{3}$NUS \qquad
$^{4}$Zhejiang Lab \qquad
$^{5}$Shanghai AI Lab \vspace{0.5mm}
\emails
{dhzhou@link.cuhk.edu.hk, huxiaowei@pjlab.org.cn} \vspace{0.5mm} \\
{\tt\small\textcolor{cvprblue}{\url{https://correr-zhou.github.io/MagicTailor}}}
}

\begin{document}

\maketitle

\blfootnote{
\hangindent=1.8em \hangafter=1
    *\ Equal contribution.\quad $\dagger$\ Corresponding author.
    }

\vspace{-3mm}

\begin{abstract}

Text-to-image diffusion models can generate high-quality images but lack fine-grained control of visual concepts, limiting their creativity.
Thus, we introduce \textbf{component-controllable personalization}, a new task that enables users to customize and reconfigure individual components within concepts. This task faces two challenges: \textit{semantic pollution}, where undesired elements disrupt the target concept, and \textit{semantic imbalance}, which causes disproportionate learning of the target concept and component.
To address these, we design \textbf{\method}, a framework that uses \textit{Dynamic Masked Degradation} to adaptively perturb unwanted visual semantics and \textit{Dual-Stream Balancing} for more balanced learning of desired visual semantics. 
The experimental results show that \method achieves superior performance in this task and enables more personalized and creative image generation.

\end{abstract}    

\section{Introduction} \label{sec:intro}

\begin{figure}[t]
    \centering
    \includegraphics[width=0.95\linewidth]{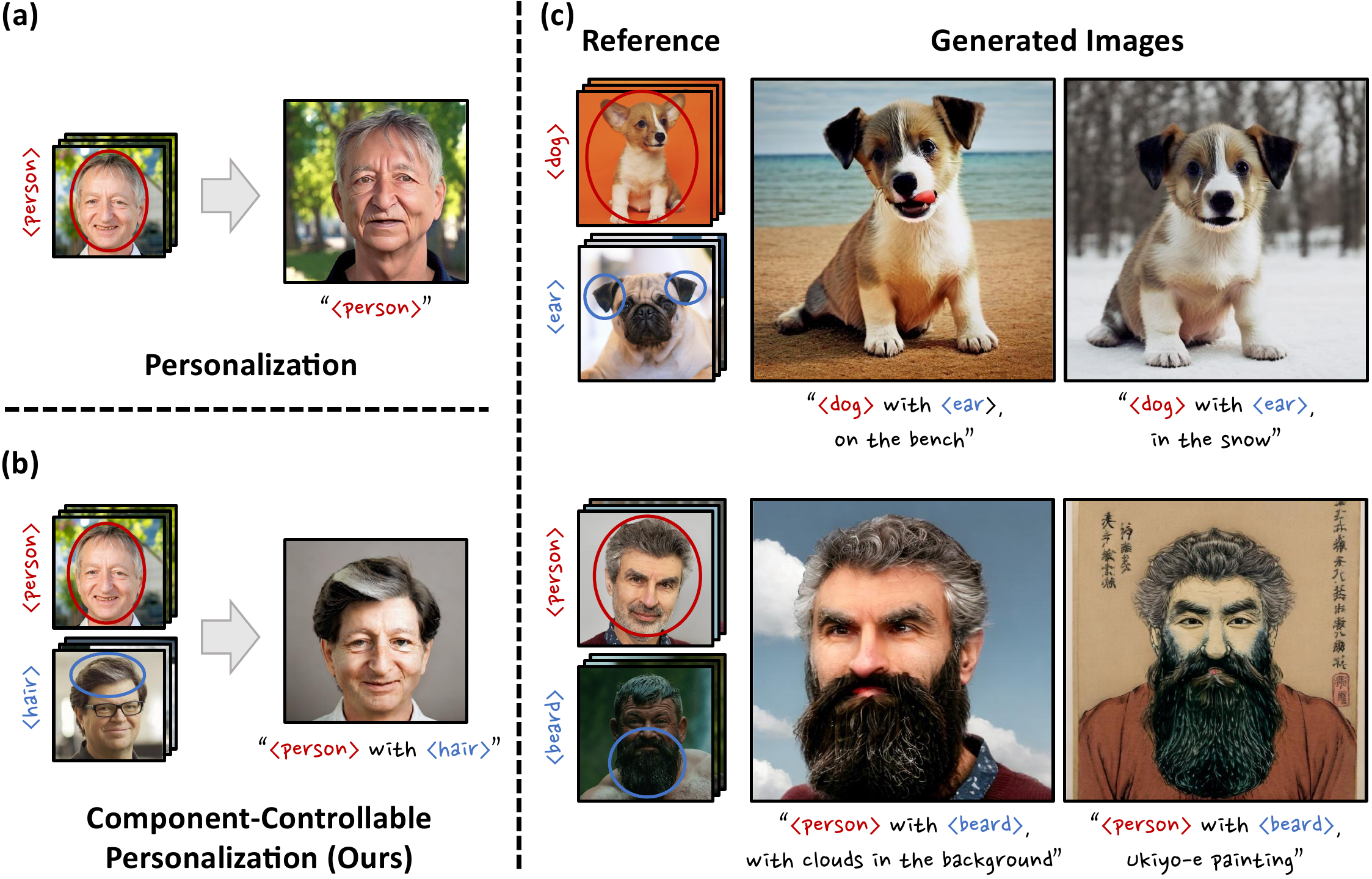}
    \vspace{-3mm}
    \captionsetup{hypcap=false}
    \captionof{figure}{ 
    \textbf{(a) Personalization:} T2I models learn from reference images and then generate predefined visual concepts. 
    \textbf{(b) Component-controllable personalization:} 
    T2I models learn from additional visual references and then enable the integration of specific components into given concepts, further unleashing creativity.
    \textbf{(c) Generated images by \method:} \method can effectively achieve component-controllable personalization.
    Note that \cred{red} and \cblue{blue} circles indicate the target concept and component, respectively. }
    \label{fig:overview}
    \vspace{-4mm}
\end{figure}

Text-to-image (T2I) diffusion models \citep{rombach2022high, ramesh2022hierarchical, chen2023pixart} have shown impressive capabilities in generating high-quality images from textual descriptions. 
While these models can generate images that align well with provided prompts, they struggle when certain visual concepts are hard to express in natural language. 
To address this, methods like \citep{gal2022image, ruiz2023dreambooth} enable T2I models to learn specific concepts from a few reference images, allowing for more accurate integration of those concepts into the generated images. This process, as shown in Fig.~\ref{fig:overview}(a), is referred as personalization.

However, existing personalization methods are limited to replicating predefined concepts and lack flexible and fine-grained control of these concepts. 
Such a limitation hinders their practical use in real-world applications, restricting their potential for creative expression. 
Inspired by the observation that concepts often comprise multiple components, a key problem in personalization lies in \textit{how to effectively control and manipulate these individual components.}

In this paper, we introduce \textbf{component-controllable personalization}, a new task that enables the reconfiguration of specific components within personalized concepts using additional visual references (Fig.~\ref{fig:overview}(b)). 
In this approach, a T2I model is fine-tuned with reference images and corresponding category labels, allowing it to learn and generate the desired concept along with the given component. 
This capability empowers users to refine and customize concepts with precise control, fostering creativity and innovation across various domains, from artworks to inventions.

\begin{figure}[t]
    \centering
    \vspace{1mm}
    \includegraphics[width=1\linewidth]{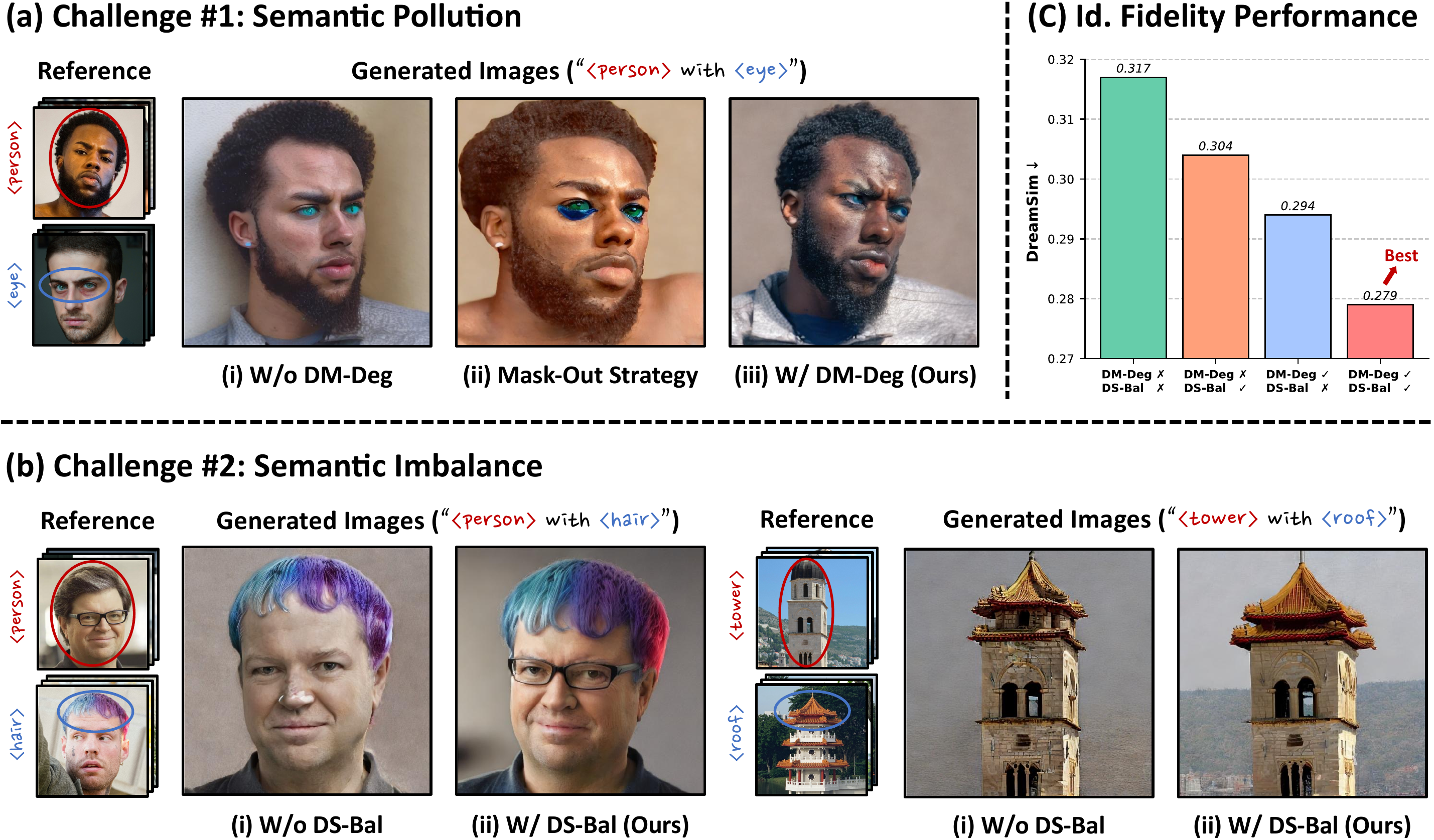}
    \vspace{-5.5mm}
    \caption{
    \textbf{Major challenges in component-controllable personalization.}
\textbf{(a) Semantic pollution:} (i) Undesired elements may interfere with the personalized concept. (ii) A simple mask-out strategy causes unintended results, while (iii) DM-Deg effectively suppresses unwanted semantics.
\textbf{(b) Semantic imbalance:} (i) Simultaneously learning the concept and component can distort either one. (ii) DS-Bal ensures balanced learning, improving personalization.
\textbf{(c) Identity fidelity performance:} Calculating DreamSim \citep{fu2023dreamsim} scores on our collected dataset, we show that DM-Deg and DS-Bal can address these challenges for faithful generation.
        }
    \label{fig:challenges}
    
    \vspace{-2.5mm}
    
\end{figure}

One challenge of this task is \emph{semantic pollution} (Fig.~\ref{fig:challenges}(a)), where unwanted visual elements inadvertently appear in generated images, ``polluting'' the personalized concept. 
This happens because the T2I model often mixes visual semantics from different regions during training. 
Masking out unwanted elements in reference images doesn’t solve the problem, as it disrupts the visual context and causes unintended compositions.
Another challenge is \emph{semantic imbalance} (Fig.~\ref{fig:challenges}(b)), where the model overemphasizes certain aspects, leading to unfaithful personalization. This occurs due to the semantic disparity between the concept and component, necessitating a more balanced learning approach to manage concept-level (\textit{e.g.}, person) and component-level (\textit{e.g.}, hair) semantics.

\begin{figure*}[t]
    \centering
    \includegraphics[width=1\linewidth]{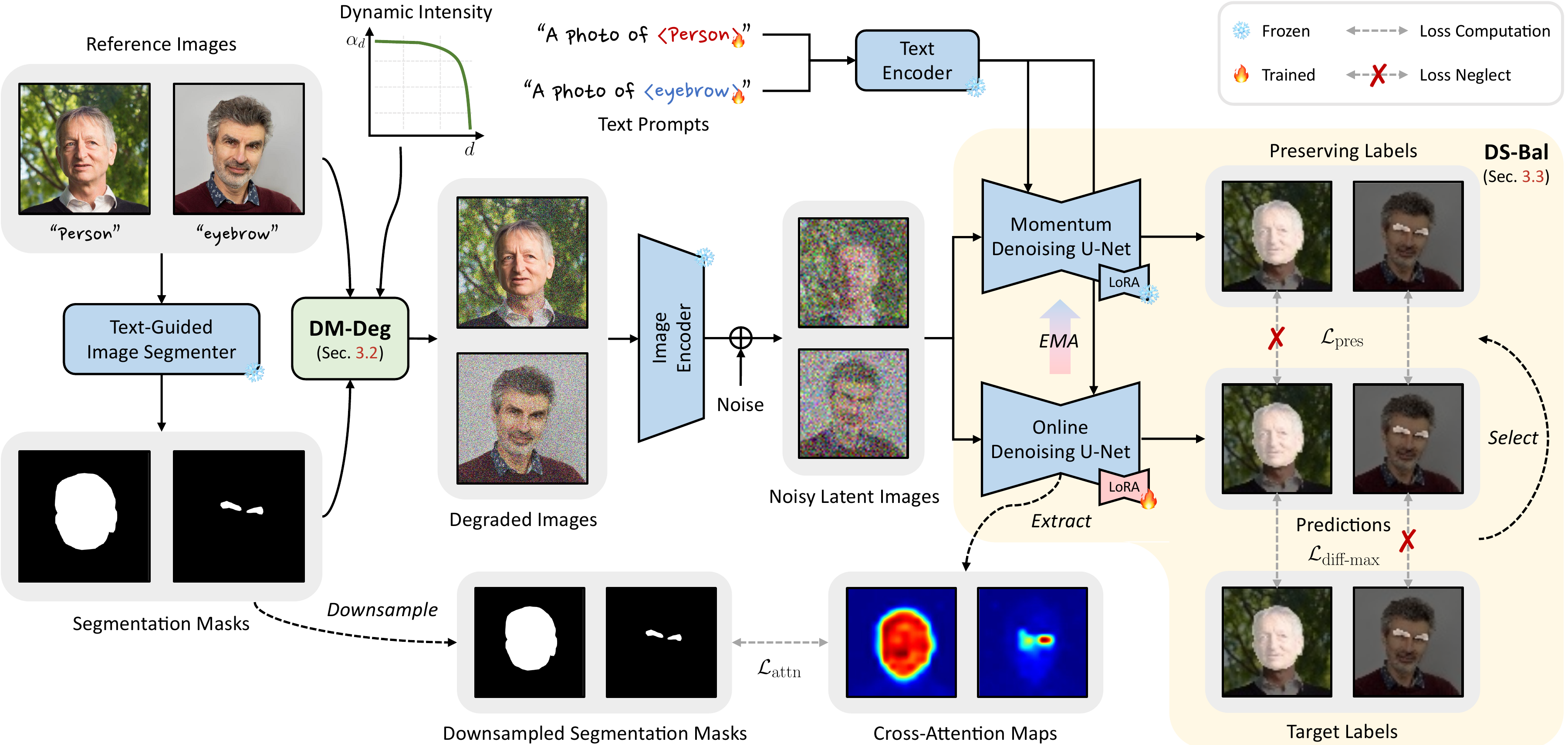}
    \vspace{-6mm}
    \caption{
    \textbf{Pipeline overview of \method.} This method fine-tunes a T2I diffusion model using reference images to learn both the target concept and component, enabling the generation of images that seamlessly integrate the component into the concept. Two key techniques, \textit{Dynamic Masked Degradation} (\textit{DM-Deg}, see Sec.~\ref{sec:deg}) and \textit{Dual-Stream Balancing} (\textit{DS-Bal}, see Sec.~\ref{sec:bal}), address \textit{semantic pollution} and \textit{semantic imbalance}, respectively. For clarity, only one image per concept/component is shown, and the warm-up stage is omitted.}
    \label{fig:pipeline}
    
    \vspace{-3mm}
    
\end{figure*}

To address these challenges, we propose \textbf{\method}, a novel framework that enables component-controllable personalization for T2I models (Fig.~\ref{fig:overview}(c)).
We first use a text-guided image segmenter to generate segmentation masks for both the concept and component and then design \emph{Dynamic Masked Degradation (DM-Deg)} to transform reference images into randomly degraded versions, perturbing undesired visual semantics. 
This method helps suppress the model’s sensitivity to irrelevant details while preserving the overall visual context, effectively mitigating \textit{semantic pollution}.
Next, we initiate a warm-up phase for the T2I model, training it on the degraded images using a masked diffusion loss to focus on the desired semantics and a cross-attention loss to strengthen the correlation between these semantics and pseudo-words.
To address \textit{semantic imbalance}, we develop \emph{Dual-Stream Balancing (DS-Bal)}, a dual-stream learning paradigm that balances the learning of visual semantics. 
In this phase, the online denoising U-Net performs sample-wise min-max optimization, while the momentum denoising U-Net applies selective preservation regularization. 
This ensures more faithful personalization of the target concept and component, resulting in outputs that better align with the intended objective.

In the experiments, we validate the superiority of \method through various qualitative and quantitative comparisons, demonstrating its state-of-the-art (SOTA) performance in component-controllable personalization. 
Moreover, detailed ablation studies and analysis further confirm the effectiveness of \method. 
In addition, we also show its potential for enabling a wide range of creative applications.

\section{Related Works} \label{sec:related_work}

\paragraph{Text-to-Image Generation.}  
T2I generation has made remarkable advancements in recent years, enabling the synthesis of vivid and diverse imagery based on textual descriptions. 
Early methods employed Generative Adversarial Networks (GANs) \citep{reed2016generative, xu2018attngan}, and transformers \citep{ding2021cogview, yu2022scaling, bai2024meissonic} also showed the potential in conditional generation. 
The advent of diffusion models has ushered in a new era in T2I generation \citep{li2024tuning, saharia2022photorealistic, ramesh2022hierarchical, chen2023pixart, xue2024raphael}. Leveraging these models, a range of related applications has rapidly emerged, including image editing \citep{li2024tuning, mou2024diffeditor, huang2024diffusion, feng2024item, huang2024dual}, image completion and translation \citep{xie2023dreaminpainter, xie2023smartbrush, lin2024difftv}, and controllable image generation \citep{zhang2023adding, wang2024diffx, zheng2023layoutdiffusion}. 
Despite advancements in T2I diffusion models, generating images that accurately reflect specific, user-defined concepts remains a challenge. 
This study explores component-controllable personalization, which allows precise adjustment of specific concepts' components using visual references.

\paragraph{Personalization.}  
Personalization seeks to adapt T2I models to generate given concepts using reference images. 
Initial approaches \citep{gal2022image, ruiz2023dreambooth} addressed this task by either optimizing text embeddings or fine-tuning the entire T2I model. 
Additionally, low-rank adaptation (LoRA) \citep{hu2021lora} has been widely adopted in this field \citep{ryu2022lora}, providing an efficient solution. 
The scope of personalization has expanded to encompass multiple concepts \citep{kumari2023multi, avrahami2023break, gu2024mix, ng2025partcraft}. 
Besides, several studies \citep{li2023photomaker, wei2023elite, zhang2024ssr, song2024moma} have explored tuning-free approaches for personalization, but these necessitate additional training on large-scale image datasets \citep{zhang2024survey}. 
In contrast, \textit{\method is a tuning-base method that requires only a few images and leverages test-time optimization to enable stable performance.}
Notably, several works \citep{huang2024parts,safaee2024clic,ng2025partcraft} have also explored how to learn and customize fine-grained elements. 
However, these methods can only combine elements or process one element at the same level.
By comparison, \textit{\method is a versatile framework able to handle both component-level and concept-level elements.}

\section{Methodology} 
\label{sec:method}

Let $\mathcal{I} = \{(\{I_{nk}\}_{k=1}^K, c_n)\}_{n=1}^N$ denote a concept-component pair with $N$ samples of concepts and components, where each sample contains $K$ reference images $\{I_{nk}\}_{k=1}^K$ with a category label $c_n$. 
In this work, we focus on a practical setting involving one concept and one component.
Specifically, we set $N = 2$ and define the first sample as a concept (\eg, dog) while the second one as a component (\eg, ear). 
In addition, these samples are associated with the pseudo-words $\mathcal{P} = \{p_n\}_{n=1}^N$ serving as their text identifiers. 
The goal of \textit{component-controllable personalization} is to fine-tune a text-to-image (T2I) model to accurately learn both the concept and the component from $\mathcal{I}$. Using text prompts with $\mathcal{P}$, the fine-tuned model should generate images that integrate the personalized concept with the specified component.

This section begins by providing an overview of the \method pipeline in Sec.~\ref{sec:pipeline} and then delves into its two core techniques in Sec.~\ref{sec:deg} and Sec.~\ref{sec:bal}.

\subsection{Overall Pipeline}
\label{sec:pipeline}

The overall pipeline of \method is illustrated in Fig.~\ref{fig:pipeline}. The process begins with identifying the desired concept or component within each reference image $I_{nk}$, employing an off-the-shelf text-guided image segmenter to generate a segmentation mask $M_{nk}$ based on $I_{nk}$ and its associated category label $c_n$. 
Conditioned on $M_{nk}$, we design \textit{Dynamic Masked Degradation (DM-Deg)} to perturb undesired visual semantics within $I_{nk}$, addressing \textit{semantic pollution}. 
At each training step, DM-Deg transforms $I_{nk}$ into a randomly degraded image $\hat{I}_{nk}$, with the degradation intensity being dynamically regulated.
Subsequently, these degraded images, along with structured text prompts, are used to fine-tune a T2I diffusion model to facilitate concept and component learning. 
The model is formally expressed as $\{\epsilon_\theta, \tau_\theta, \mathcal{E}, \mathcal{D}\}$, where $\epsilon_\theta$ represents the denoising U-Net, $\tau_\theta$ is the text encoder, and $\mathcal{E}$ and $\mathcal{D}$ denote the image encoder and decoder, respectively.
To promote the learning of the desired visual semantics, we employ the masked diffusion loss, which is defined as:
\begin{equation}
    \mathcal{L}_{\text{diff}} \ = \ 
    \mathbb{E}_{n, k, \epsilon, t}
    \Big [
    \big \Vert 
    \epsilon_n \odot M_{nk}^\prime - 
    \epsilon_\theta(z_{nk}^{(t)}, t, e_n) \odot M_{nk}^\prime
    \big \Vert_2^2
    \Big ] \ ,
    \label{eq:diff}
\end{equation}
where $\epsilon_n \sim \mathcal{N}(0, 1)$ is the unscaled noise, $z_{nk}^{(t)}$ is the noisy latent image of $\hat{I}_{nk}$ with a random time step $t$,
$e_n$ is the text embedding of the corresponding text prompt,
and $M_{nk}^\prime$ is downsampled from $M_{nk}$ to match the shape of $\epsilon$ and $z_{nk}$.
Additionally, we incorporate the cross-attention loss to strengthen the correlation between desired visual semantics and their corresponding pseudo-words, formulated as:
\begin{equation}
    \mathcal{L}_{\text{attn}} \ = \
    \mathbb{E}_{n, k, t}
    \Big [
    \big \Vert 
    A_\theta (p_n, z_{nk}^{(t)}) - M_{nk}^{\prime \prime}
    \big \Vert_2^2
    \Big ] \ ,
    \label{eq:attn}
\end{equation}
when $A_\theta (p_n, z_{nk}^{(t)})$ is the cross-attention maps between the pseudo-word $p_n$ and the noisy latent image $z_{nk}^{(t)}$ and $M_{nk}^{\prime \prime}$ is downsampled from $M_{nk}$ to match the shape of $A_\theta (p_n, z_{nk}^{(t)})$.

Using $\mathcal{L}_{\text{diff}}$ and $\mathcal{L}_{\text{attn}}$, we first warm up the T2I model by jointly learning all samples, aiming to preliminarily inject the knowledge of visual semantics.
The loss of the warm-up stage is defined as:
\begin{equation}
    \mathcal{L}_{\text{warm-up}} \ = \ 
    \mathcal{L}_{\text{diff}} \ + \
    \lambda_{\text{attn}}  \mathcal{L}_{\text{attn}} \ ,
    \label{eq:warmup}
\end{equation}
where $\lambda_{\text{attn}}=0.01$ is the loss weight for $\mathcal{L}_{\text{attn}}$.
For efficient fine-tuning, we only train the denoising U-Net $\epsilon_\theta$ in a low-rank adaptation (LoRA) \citep{hu2021lora} manner and the text embedding of the pseudo-words $\mathcal{P}$, keeping the others frozen.
Thereafter, we employ \textit{Dual-Stream Balancing (DS-Bal)} to address \textit{semantic imbalance}.  
In this paradigm, the online denoising U-Net $\epsilon_\theta$ conducts sample-wise min-max optimization for the hardest-to-learn sample, and meanwhile the momentum denoising U-Net $\tilde{\epsilon}_\theta$ applies selective preserving regularization for the other samples.

\subsection{Dynamic Masked Degradation}
\label{sec:deg}

\textit{Semantic pollution} is a significant challenge for component-controllable personalization.
As shown in Fig.~\ref{fig:challenges}(a.i), 
the target concept (\ie, person) can be distorted by the owner of the target component (\ie, eye), resulting in a hybrid person. Masking regions outside the target concept and component can damage the overall context, leading to overfitting and odd compositions (Fig.~\ref{fig:challenges}(a.ii)). 
To address this, undesired visual semantics in reference images must be handled appropriately. We propose \textit{Dynamic Masked Degradation (DM-Deg)}, which dynamically perturbs undesired semantics to suppress their influence on the T2I model while preserving the overall visual context (Fig.~\ref{fig:challenges}(a.iii)\&(c)).

\begin{figure}

    \centering
    \includegraphics[width=0.95\linewidth]{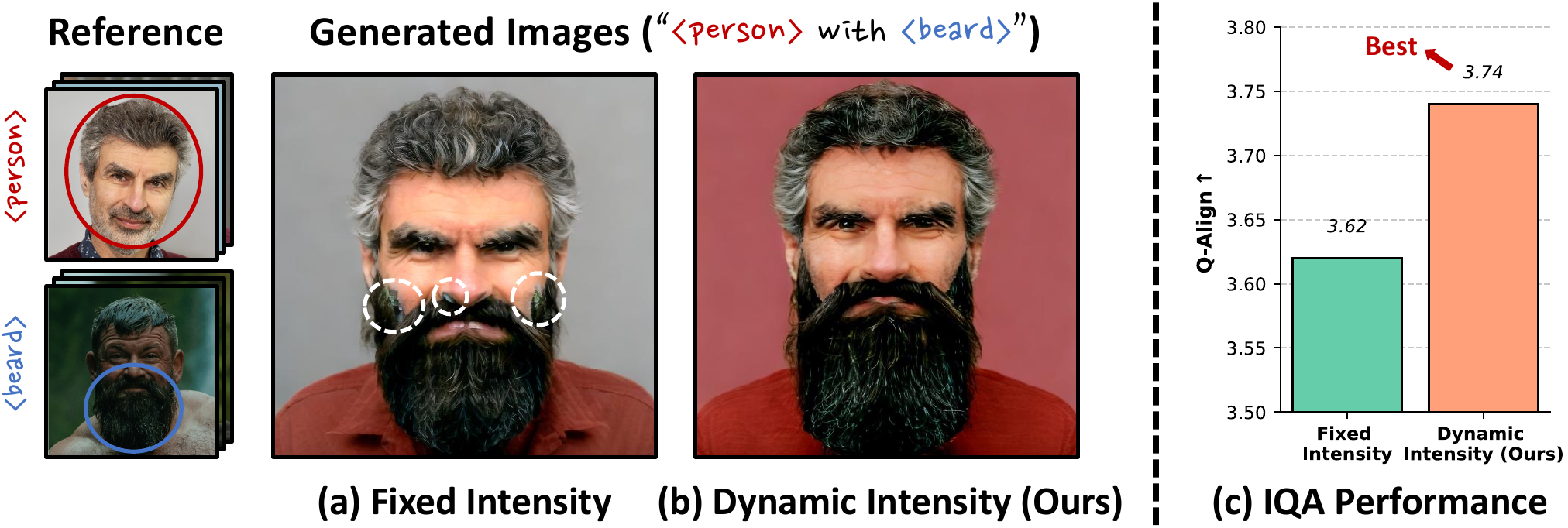}
    \vspace{-2mm}
    \caption
    {
        \textbf{Motivation of dynamic intensity.}
        (a) Fixed intensity ($\alpha_d = 0.5$ here) could cause noisy generated images.
        (b) Our dynamic intensity can mitigate noise memorization.
        (c) We report IQA results of Q-Align \citep{wu2023q} on our dataset, showing that our dynamic intensity helps to enhance the quality of generated images.
    }
    \label{fig:dyn}
    \vspace{-3mm}   
\end{figure}

\paragraph{Degradation Imposition.}
In each training step, DM-Deg imposes degradation in the out-of-mask region for each reference image.
We use Gaussian noise for degradation due to its simplicity. For a reference image $I_{nk}$, we randomly sample a Gaussian noise matrix $G_{nk} \sim \mathcal{N}(0, 1)$ with the same shape as $I_{nk}$, where the pixel values of $I_{nk}$ range from $-1$ to $1$. The degradation is then applied as follows:
\begin{equation}
\hat{I}_{nk} = \alpha_d G_{nk} \odot (1 - M_{nk}) + I_{nk},
\end{equation}
where $\odot$ denotes element-wise multiplication, and $\alpha_d \in [0, 1]$ is a dynamic weight controlling the degradation intensity. 
While previous works \citep{xiao2023fastcomposer, li2023photomaker} have used noise to fully cover the background or enhance data diversity,
DM-Deg aims to produce a degraded image $\hat{I}_{nk}$ that retains the original visual context. By introducing $\hat{I}_{nk}$, we can suppress the T2I model from perceiving undesired visual semantics in out-of-mask regions, as these semantics are perturbed by random noise at each training step.

\paragraph{Dynamic Intensity.}
Unfortunately, the T2I model may gradually memorize the introduced noise while learning meaningful visual semantics, leading to noise appearing in generated images (Fig.~\ref{fig:dyn}(a)).
This behavior is consistent with previous observations on deep networks \citep{arpit2017closer}.
To address this, we propose a descending scheme that dynamically regulates the intensity of the imposed noise during training. 
 This scheme follows an exponential curve, maintaining a relatively high intensity in the early stages and decreasing it sharply in later stages.
Let $d$ denote the current training step and $D$ denote the total training step. The curve of dynamic intensity is defined as:
\begin{equation}
    \alpha_d  \ = \ \alpha_{\text{init}} (1 - (\frac{d}{D})^{\gamma}) \ ,
    \label{eq:dyn_int}
\end{equation}
where $\alpha_{\text{init}}$ is the initial value of $\alpha_d$ and $\gamma$ controls the descent rate.
We empirically set $\alpha_{\text{init}} = 0.5$ and $\gamma = 32$, tuned within the powers of $2$.
This dynamic intensity scheme effectively prevents semantic pollution and significantly mitigates the memorization of introduced noise, leading to improved generation performance (Fig.~\ref{fig:dyn}(b)\&(c)).

\begin{figure}[t]
    \centering
    \includegraphics[width=1\linewidth]{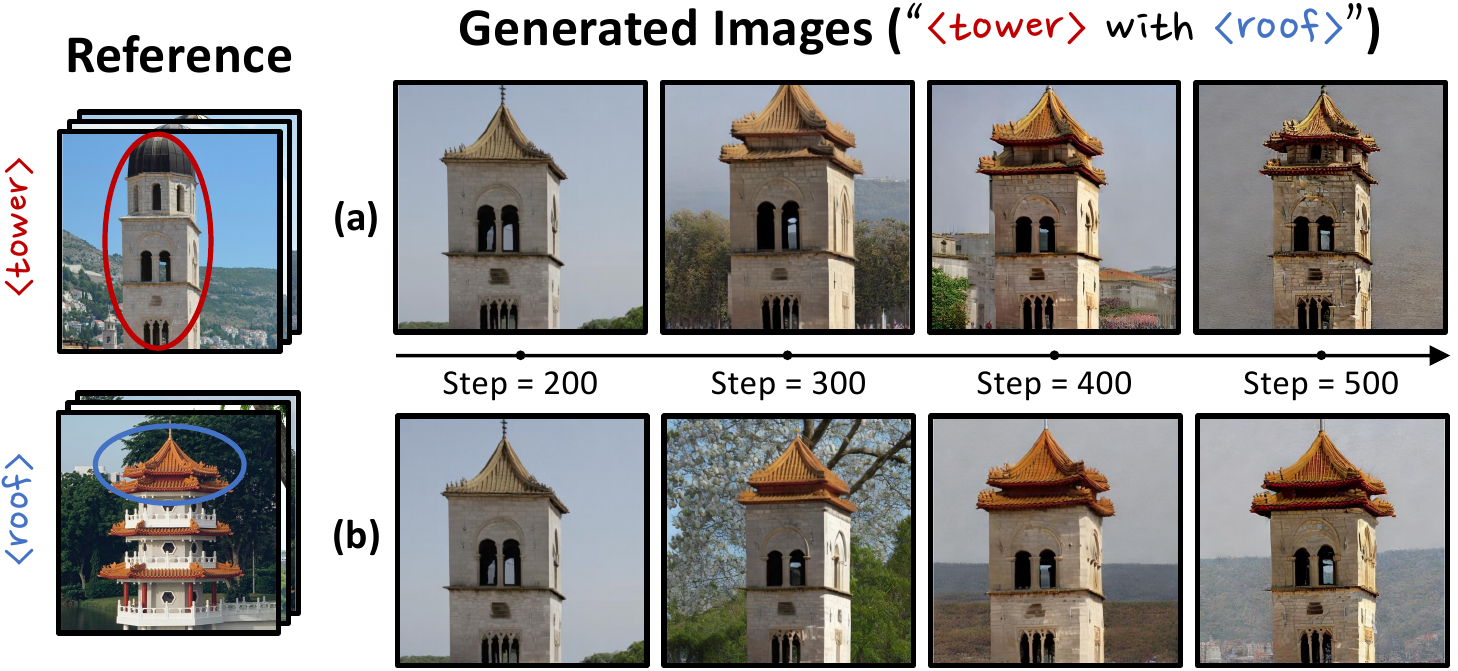}
    
    \vspace{-2mm}

    \caption
    {
        \textbf{Learning process visualization.}
        (a) The vanilla learning paradigm tends to overemphasize the easier one.
        (b) DS-Bal effectively balances the learning of the concept and component.
    }
    \label{fig:imb}
    \vspace{-3mm}  
    
\end{figure}

\subsection{Dual-Stream Balancing}
\label{sec:bal}

Another key challenge is \textit{semantic imbalance}, which arises from the disparity in visual semantics between the target concept and its component.
Specifically, concepts generally possess richer visual semantics than components (\eg, person vs. hair),
but in some cases, components may have more complex semantics (\eg, simple tower vs. intricate roof).
This imbalance complicates joint learning, leading to overemphasis on either the concept or the component, and resulting in incoherent generation (Fig.~\ref{fig:imb}(a)). 
To address this, we design \textit{Dual-Stream Balancing (DS-Bal)},  a dual-stream learning paradigm integrated with online and momentum denoising U-Nets (Fig.~\ref{fig:pipeline}) for balanced semantic learning, aiming to improve personalization fidelity (Fig.~\ref{fig:imb}(b) \& Fig.~\ref{fig:challenges}(c)). 

\paragraph{Sample-Wise Min-Max Optimization.} 
From a loss perspective, the visual semantics of the concept and component are learned by optimizing the masked diffusion loss $\mathcal{L}_{\text{diff}}$ across all the samples.
However, this indiscriminate optimization fails to allocate sufficient learning effort to a more challenging sample, leading to an imbalanced learning process.
To address this, DS-Bal uses the online denoising U-Net to focus on learning the hardest-to-learn sample at each training step.
Inheriting the weights of the original denoising U-Net, which is warmed up through joint learning, the online denoising U-Net $\epsilon_\theta$ optimizes only the sample with the highest masked diffusion loss as:
\begin{align}
    \mathcal{L}_{\text{diff-max}}  \ = \ 
    \max_n
    \mathbb{E}_{k, \epsilon, t}
    \Big [
    \big \Vert 
    & \epsilon_n \odot M_{nk}^\prime - \notag \\
    & \epsilon_\theta(z_{nk}^{(t)}, t, e_n) \odot M_{nk}^\prime
    \big \Vert_2^2
    \Big ] \ ,
    \label{eq:diff_max}
\end{align}
where minimizing $\mathcal{L}_{\text{diff-max}}$ can be considered as a form of min-max optimization \citep{razaviyayn2020nonconvex}.
The learning objective of $\epsilon_\theta$ may switch across different training steps and is not consistently dominated by the concept or component.
Such an optimization scheme can effectively modulate the learning dynamics of multiple samples and avoid the overemphasis on any particular one.

\begin{figure}[t]

    \centering

    \includegraphics[width=1\linewidth]{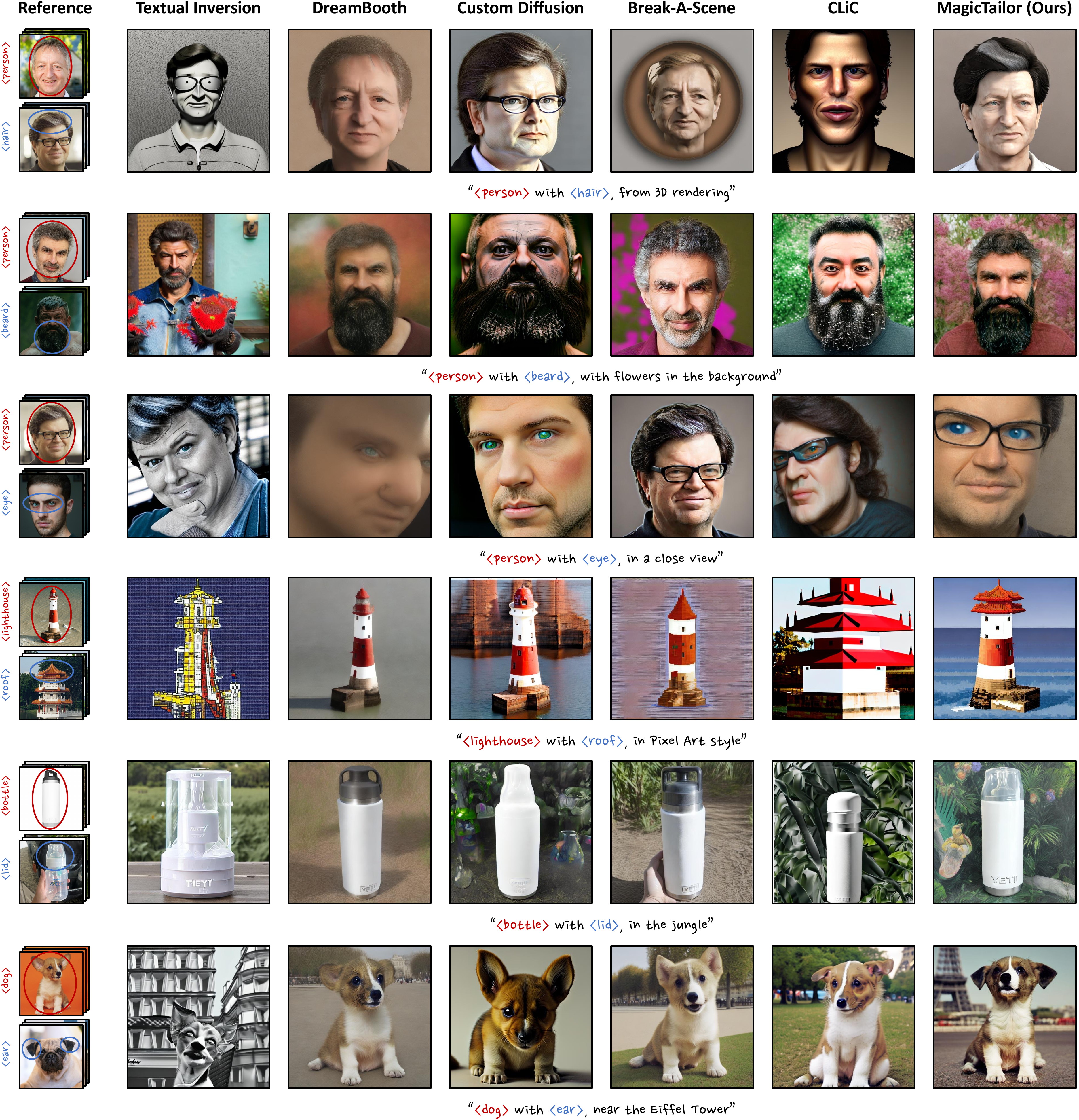}
    \vspace{-6mm}
    \caption
    {
        \textbf{Qualitative comparisons.} 
    We present images generated by \method and other methods across various domains. \method achieves better text alignment, identity fidelity, and generation quality. 
    \textit{Due to space limitations, please zoom in for a better view}. 
    More results are provided in Appendix~\ref{app:app_more_qual}.
    }
    
    \label{fig:qual_results}
    
    \vspace{-2mm}
     
\end{figure}

\paragraph{Selective Preserving Regularization.}
At a training step, the sample neglected in $\mathcal{L}_{\text{diff-max}}$ may suffer from knowledge forgetting.
This is because the optimization of $\mathcal{L}_{\text{diff-max}}$, which aims to enhance the knowledge of a specific sample, could inadvertently overshadow the knowledge of the others.
In light of this, DS-Bal meanwhile exploits the momentum denoising U-Net $\tilde{\epsilon}_\theta$ to preserve the learned visual semantics of the other sample in each training step. 
Specifically, we first select the sample that is excluded in $\mathcal{L}_{\text{diff-max}}$, which is expressed as
$
    S = \{n | n = 1, ..., N\} - \{n_\text{max} \},
$
where $n_\text{max}$ is the index of the target sample in $\mathcal{L}_{\text{diff-max}}$ and $S$ is the selected index set.
Then, we use $\tilde{\epsilon}_\theta$ to apply regularization for $S$,
with the masked preserving loss as:
\begin{align}
    \mathcal{L}_{\text{pres}} \ = \ 
    \mathbb{E}_{n \in S, k, t}
    \Big [
    \big \Vert 
    \tilde{\epsilon}_\theta(z_{nk}^{(t)}, t, e_n) & \odot M_{nk}^\prime - \notag \\
    \epsilon_\theta(z_{nk}^{(t)}, t, e_n) & \odot M_{nk}^\prime
    \big \Vert_2^2
    \Big ] \ ,
    \label{eq:pres}
\end{align}
where $\tilde{\epsilon}_\theta$ is updated from $\epsilon_\theta$ using EMA \citep{tarvainen2017mean} with the smoothing coefficient $\beta = 0.99$, thereby sustaining the prior accumulated knowledge of $\epsilon_\theta$ in each training step.
By encouraging the consistency between the output of $\epsilon_\theta$ and $\tilde{\epsilon}_\theta$ in $\mathcal{L}_{\text{pres}}$, we can facilitate the knowledge maintenance of the other samples while learning a specific sample in $\mathcal{L}_{\text{diff-max}}$.
Overall, DS-Bal can be considered a mechanism to adaptively assign target labels $\epsilon_n$ or preserving labels $\tilde{\epsilon}_\theta(z_{nk}^{(t)}, t, e_n)$ to different samples, enabling dynamic loss supervision (Fig.~\ref{fig:pipeline}).
Using a loss weight $\lambda_{\text{pres}} = 0.2$, the total loss of the DS-Bal stage is formulated as:
\begin{equation}
    \mathcal{L}_{\text{DS-Bal}} \ = \ 
    \mathcal{L}_{\text{diff-max}} \ + \
    \lambda_{\text{pres}} \mathcal{L}_{\text{pres}}  \ + \
    \lambda_{\text{attn}} \mathcal{L}_{\text{attn}} \ .
    \label{eq:total_loss}
\end{equation}


\section{Experimental Results} \label{sec:exps}

\begin{table}[t]
    
    \centering
        \caption
     {
         \textbf{Quantitative comparisons on automatic metrics.}
         \method can achieve SOTA performance on all four automatic metrics.
        The best results are marked in bold.
     }
    \vspace{-2mm}
    \setlength{\tabcolsep}{1.1mm}  
    \renewcommand{\arraystretch}{1.3}
    
    \resizebox{\linewidth}{!}
    {
        \begin{tabular}{lcccc}
        \specialrule{0.12em}{0pt}{0pt}
        Methods & CLIP-T $\uparrow$ & CLIP-I $\uparrow$ & DINO $\uparrow$ & DreamSim $\downarrow$ \\
        \midrule
        Textual Inversion \citep{gal2022image} & 0.236  & 0.742  & 0.620  & 0.558  \\
        DreamBooth \citep{ruiz2023dreambooth} & 0.266  & 0.841  & 0.798  & 0.323  \\
        Custom Diffusion \citep{kumari2023multi} & 0.251  & 0.797  & 0.750  & 0.407  \\
        Break-A-Scene \citep{avrahami2023break} & 0.259  & 0.840  & 0.780  & 0.338  \\
        CLiC \citep{safaee2024clic} & 0.263  & 0.764  & 0.663  & 0.499  \\
        MagicTailor (Ours) & \textbf{0.270} & \textbf{0.854} & \textbf{0.813} & \textbf{0.279} \\
        \specialrule{0.12em}{0pt}{0pt}
        \end{tabular}%
    }    

    \vspace{-1mm}

    \label{tab:auto_metrics}
    
\end{table}

\begin{table}[t]
    
    \centering
       
    \caption
     {
         \textbf{Quantitative comparisons on the user study.}
         \method also outperforms other methods in all aspects of human evaluation.
     }
    \vspace{-2mm}
    \setlength{\tabcolsep}{1.2mm}  
    \renewcommand{\arraystretch}{1.3}
    
    \resizebox{\linewidth}{!}
    {
        \begin{tabular}{lccc}
        \specialrule{0.12em}{0pt}{0pt}
        Methods & {Text Align. $\uparrow$} & {Id. Fidelity $\uparrow$} & {Gen. Quality $\uparrow$} \\
        \midrule
        Textual Inversion \citep{gal2022image} & 5.8\% & 2.5\% & 5.2\% \\
        DreamBooth \citep{ruiz2023dreambooth} & 15.3\% & 14.7\% & 12.5\% \\
        Custom Diffusion \citep{kumari2023multi} & 7.1\% & 7.7\% & 9.8\% \\
        Break-A-Scene \citep{avrahami2023break} & 10.8\% & 12.1\% & 22.8\% \\
        CLiC \citep{safaee2024clic} & 4.5\% & 5.1\% & 6.2\% \\
        MagicTailor (Ours) & \textbf{56.5\%} & \textbf{57.9\%} & \textbf{43.4\%} \\
        \specialrule{0.12em}{0pt}{0pt}
        \end{tabular}%
    }

    \vspace{-1mm}
    
    \label{tab:user_study}
    
\end{table}

\subsection{Experimental Setup}

\paragraph{Dataset, Implementation, and Evaluation.}  
For a systematic investigation, we collect a dataset from diverse domains, including characters, animation, buildings, objects, and animals. We use Stable Diffusion (SD) 2.1 \citep{rombach2022high} as the pretrained T2I model.
For the warm-up and DS-Bal stages, we set the training steps to 200 and 300, with learning rates of \( 1 \times 10^{-4} \) and \( 1 \times 10^{-5} \), respectively. Each concept-component pair requires only about five minutes of training on an A100 GPU. 
For evaluation, we design 20 text prompts covering a wide range of scenarios and generate 14,720 images for each method. 
To ensure fairness, all random seeds are fixed during both training and inference.
More details of the experimental setup are included in Appendix~\ref{app:setup_details}.

\paragraph{Compared Methods.}
We compare our \method with several personalization methods, including Textual Inversion (TI) \citep{gal2022image}, DreamBooth (DB) \citep{ruiz2023dreambooth}, Custom Diffusion (CD) \citep{kumari2023multi}, Break-A-Scene (BAS) \citep{avrahami2023break}, and CLiC \citep{safaee2024clic}. These methods were selected for their representativeness of personalization frameworks or relevance to learning fine-grained elements. For a fair comparison, we adapt them to our task with minimal modifications, specifically by incorporating the masked diffusion loss (Eq.~\ref{eq:diff}).
Apart from method-specific configurations, all methods are implemented using the same setup to ensure consistency.

\subsection{Qualitative Comparisons}

The qualitative results are shown in in Fig.~\ref{fig:qual_results}.
As observed, TI, CD, and CLiC primarily suffer from semantic pollution, where undesired visual semantics significantly distort the personalized concept. 
Besides, DB and BAS also struggle in this challenging task, with an overemphasis on either the concept or the component due to semantic imbalance, sometimes even causing the target component to be completely absent. 
An interesting finding is that imbalanced learning can exacerbate semantic pollution, leading to the color and texture of the target concept or component being mistakenly transferred to unintended parts of the generated images. 
In contrast, \method effectively generates text-aligned images that accurately represent both the target concept and component. 
To further demonstrate the performance of \method, we provide additional comparisons in Appendix~\ref{app:app_add_comp}.

\begin{table}[t]
    
    \centering
        
    \caption
    {
        \textbf{Effectiveness of key techniques.}
        Our DM-Deg and DS-Bal effectively contribute to a superior performance trade-off.
    }
    \vspace{-2mm}
    \setlength{\tabcolsep}{3.4mm}  
    \renewcommand{\arraystretch}{1.3} 
        
    \resizebox{\linewidth}{!}
    {
        \begin{tabular}{cccccc}
            \specialrule{0.12em}{0pt}{0pt}
            DM-Deg & DS-Bal & CLIP-T $\uparrow$ & CLIP-I $\uparrow$ & DINO $\uparrow$ & DreamSim $\downarrow$ \\
            \midrule
                  &       & 0.275 & 0.837 & 0.798 & 0.317 \\
            $\checkmark$ &       & \textbf{0.276} & 0.848 & 0.809 & 0.294 \\
                  & $\checkmark$ & 0.270 & 0.845 & 0.802 & 0.304 \\
            $\checkmark$ & $\checkmark$ & 0.270 & \textbf{0.854} & \textbf{0.813} & \textbf{0.279} \\
            \specialrule{0.12em}{0pt}{0pt}
        \end{tabular}%
    }

    \vspace{-2mm}  
    
    \label{tab:ablation_tech}
    
\end{table}
\begin{figure}[t]

    \centering
    
    \includegraphics[width=0.9\linewidth]{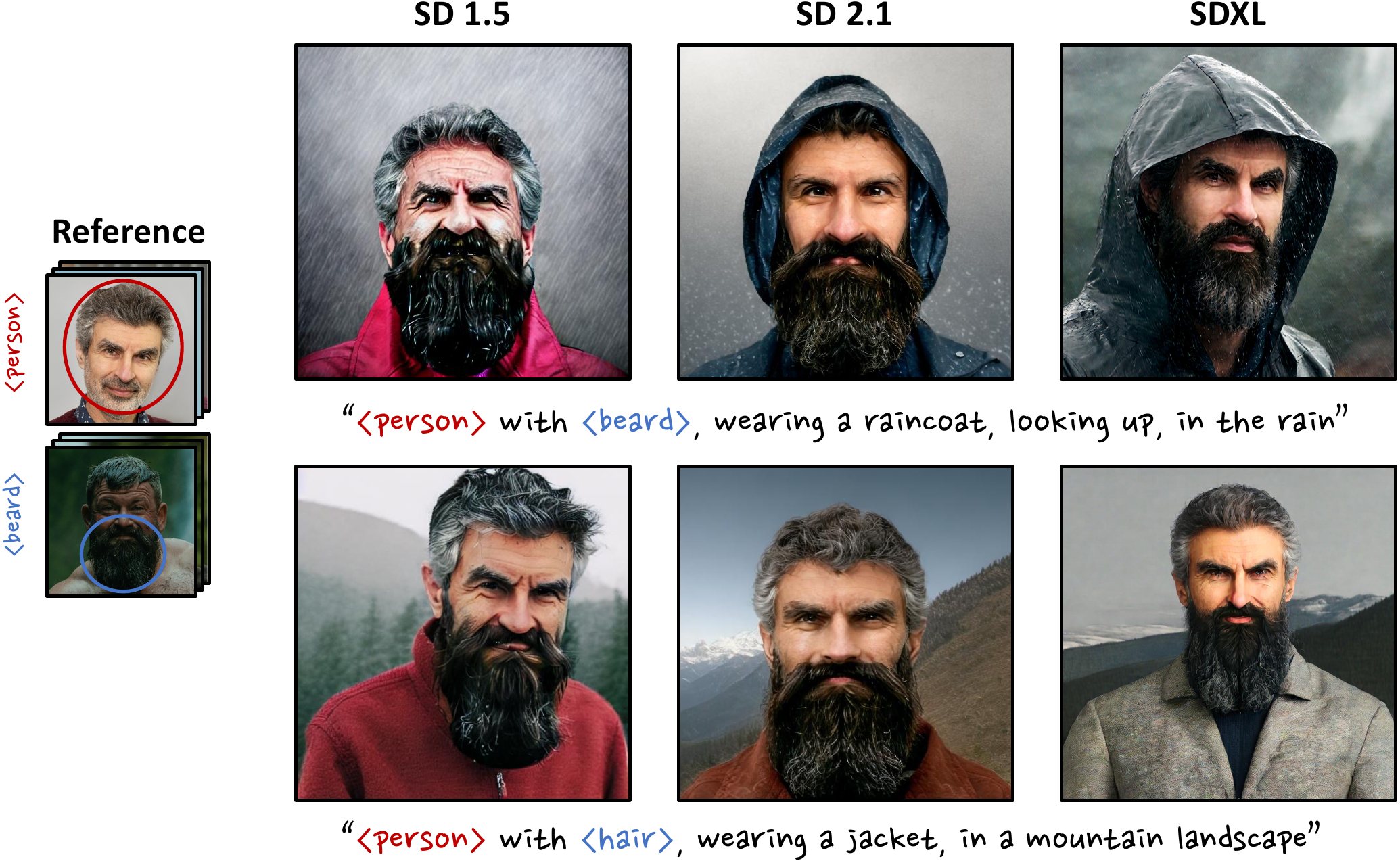}
    
    \vspace{-2mm}
    
    \caption
    {
        \textbf{Compatibility with different backbones.}
        We equip \method with SD 1.5 \citep{rombach2022high}, SD 2.1 \citep{rombach2022high}, and SDXL \citep{podell2023sdxl}. The results show that \method can be generalized to multiple backbones, and a better backbone could provide better generation quality.
    }
    
    \vspace{-3mm}
    
    \label{fig:diff_backbone}

\end{figure}

\begin{figure}[t]

    \vspace{-1mm}
    
    \centering
    
    \includegraphics[width=1\linewidth]{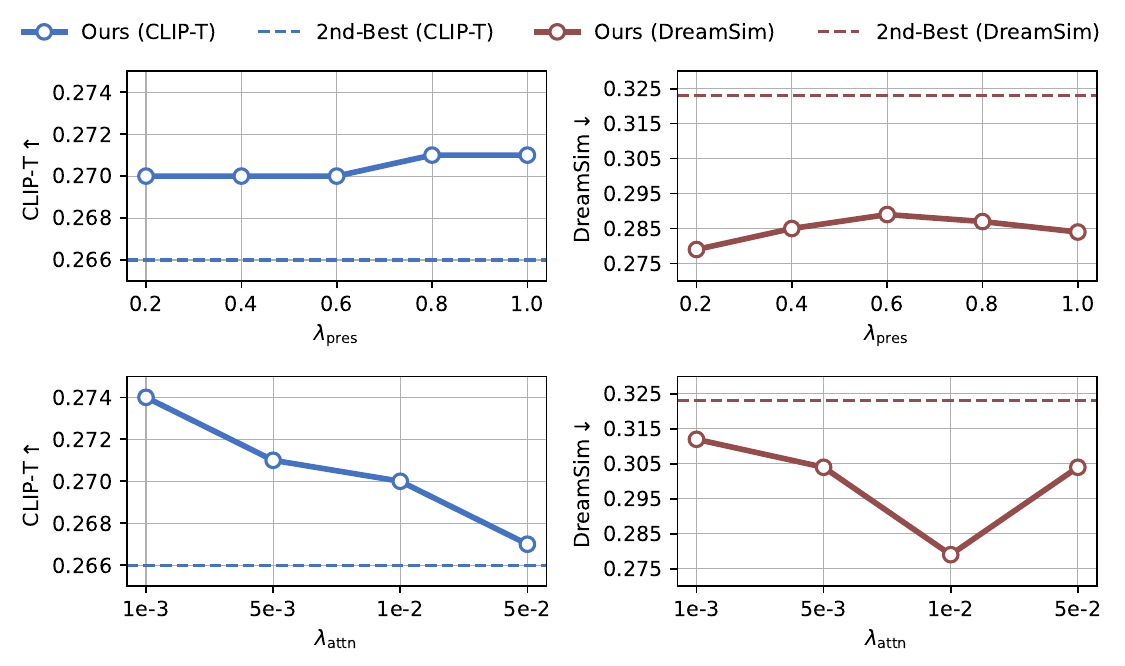}

    \vspace{-4mm}
    
    \caption
    {
        \textbf{Robustness on loss weights.}
        We report CLIP-T  \citep{gal2022image} for text alignment, and DreamSim \citep{fu2023dreamsim} for identity fidelity as it is most similar to human judgments.
        Second-best results in Table~\ref{tab:auto_metrics} are also presented to highlight our robustness.
    }
    
    \label{fig:loss_weights}
    
    \vspace{-3mm}
     
\end{figure}
\begin{figure}[t]

    \centering
    
    \includegraphics[width=\linewidth]{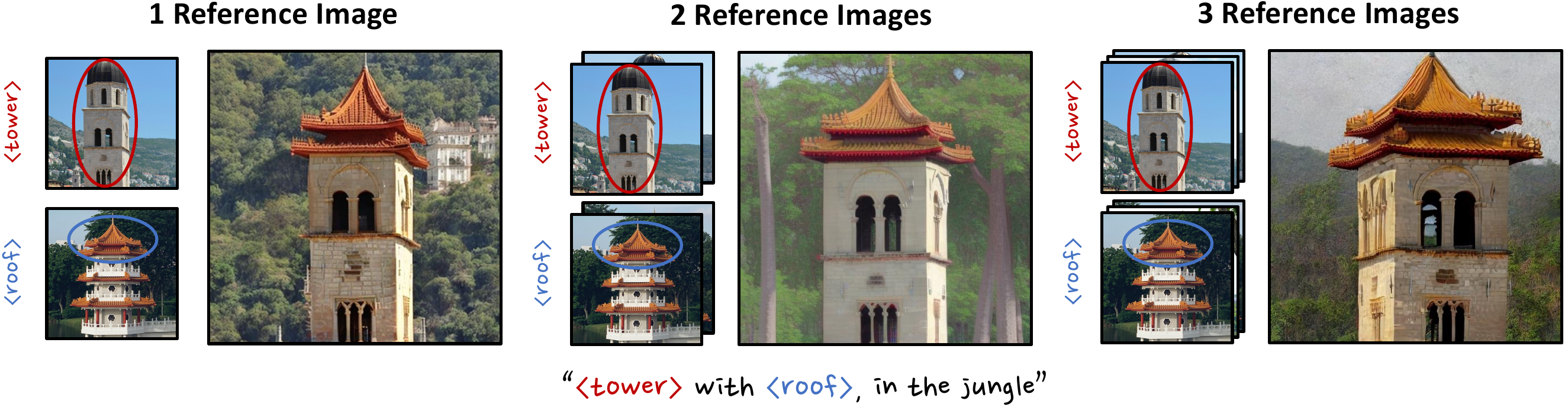}
    
    \vspace{-3mm}
    
    \caption
    {
        \textbf{Performance on different numbers of reference images.} 
        We present qualitative results to show that \method can still achieve satisfactory performance when provided only 1 or 2 reference image(s) per concept and component.c
    }
    
    \label{fig:less_ref}
    
    \vspace{-3mm}
     
\end{figure}

\subsection{Quantitative Comparisons}
\textbf{Automatic Metrics.}
We utilize four automatic metrics in the aspects of text alignment (CLIP-T \citep{gal2022image}) and identity fidelity (CLIP-I \citep{radford2021learning}, DINO \citep{oquab2023dinov2}, DreamSim \citep{fu2023dreamsim}).
\textit{To precisely measure identity fidelity}, we segment out the concept and component in each reference and evaluation image, and then eliminate the target component from the segmented concept.
As we can see in Tab.~\ref{tab:auto_metrics}, component-controllable personalization remains a tough task even for SOTA methods of personalization. 
By comparison, \method achieves the best results in both identity fidelity and text alignment.
It should be credited to the effective framework tailored to this special task.
%

\paragraph{User Study.}
We further evaluate the methods with a user study.
Specifically, a detailed questionnaire is designed to display 20 groups of evaluation images with the corresponding text prompt and reference images.
Users are asked to select the best result in each group for three aspects, including text alignment, identity fidelity, and generation quality.
Finally, we collect a total of 3,180 valid answers and report the selected rates in Tab.~\ref{tab:user_study}.
It can be observed that \method can also achieve superior performance in human preferences, further verifying its effectiveness.

\subsection{Ablation Studies and Analysis}
We conduct comprehensive ablation studies and analysis for \method to verify its capability.
More ablation studies and analysis are included in Appendix~\ref{app:app_add_abl}.

\paragraph{Effectiveness of Key Techniques.}
In Tab.~\ref{tab:ablation_tech}, we investigate two key techniques by starting from a baseline framework described in Sec.~\ref{sec:pipeline}.
Even without DM-Deg and DS-Bal, such a baseline framework can still have competitive performance, showing its reliability.
On top of that, we introduce DM-Deg and DS-Bal, where the superior performance trade-off indicates their significance.
Qualitative results can refer to Fig.~\ref{fig:challenges}.

\begin{figure*}[t]

    \centering
    \includegraphics[width=1\linewidth]{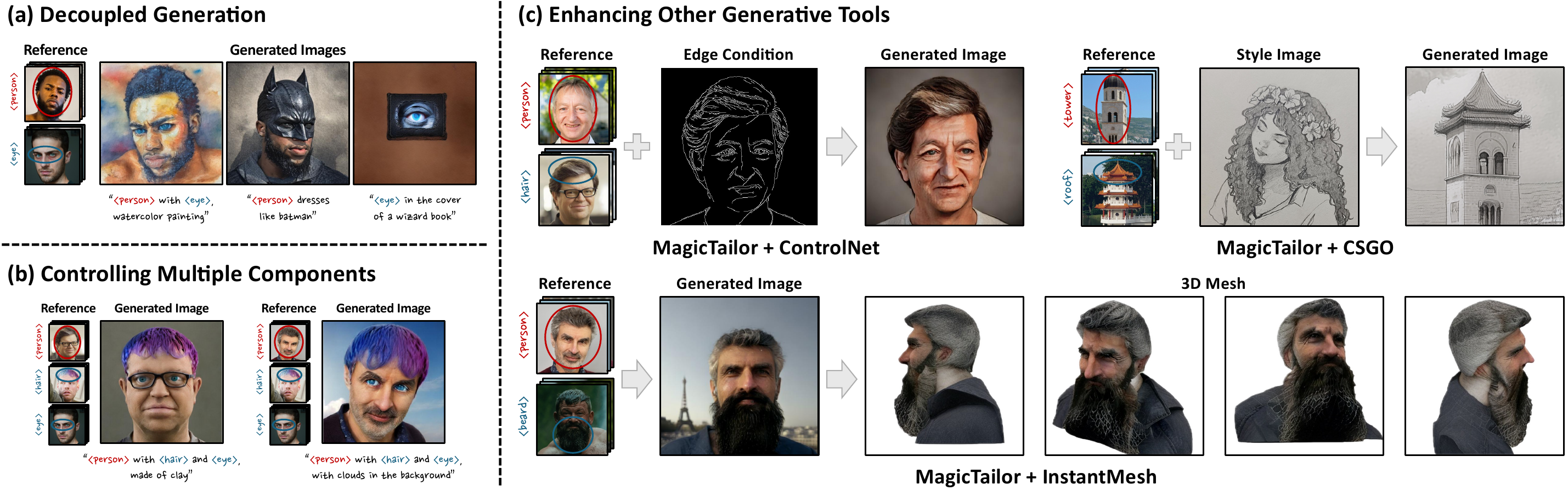}
    \vspace{-7mm}
    \caption
    {
        \textbf{Further applications of \method.}
        \textbf{(a) Decoupled generation:}  
        \method can also separately generate the target concept and component, enriching prospective combinations.
        \textbf{(b) Controlling multiple components:}  
        \method shows the potential to handle more than one component, highlighting its effectiveness.  
        \textbf{(c) Enhancing other generative tools:} 
        \method can seamlessly integrate with various generative tools, adding the capability to control components within their generation pipelines.
    }
    \label{fig:applications}
    
    \vspace{-4mm}
     
\end{figure*}

\begin{figure}[t]

    \centering
    
    \includegraphics[width=0.9\linewidth]{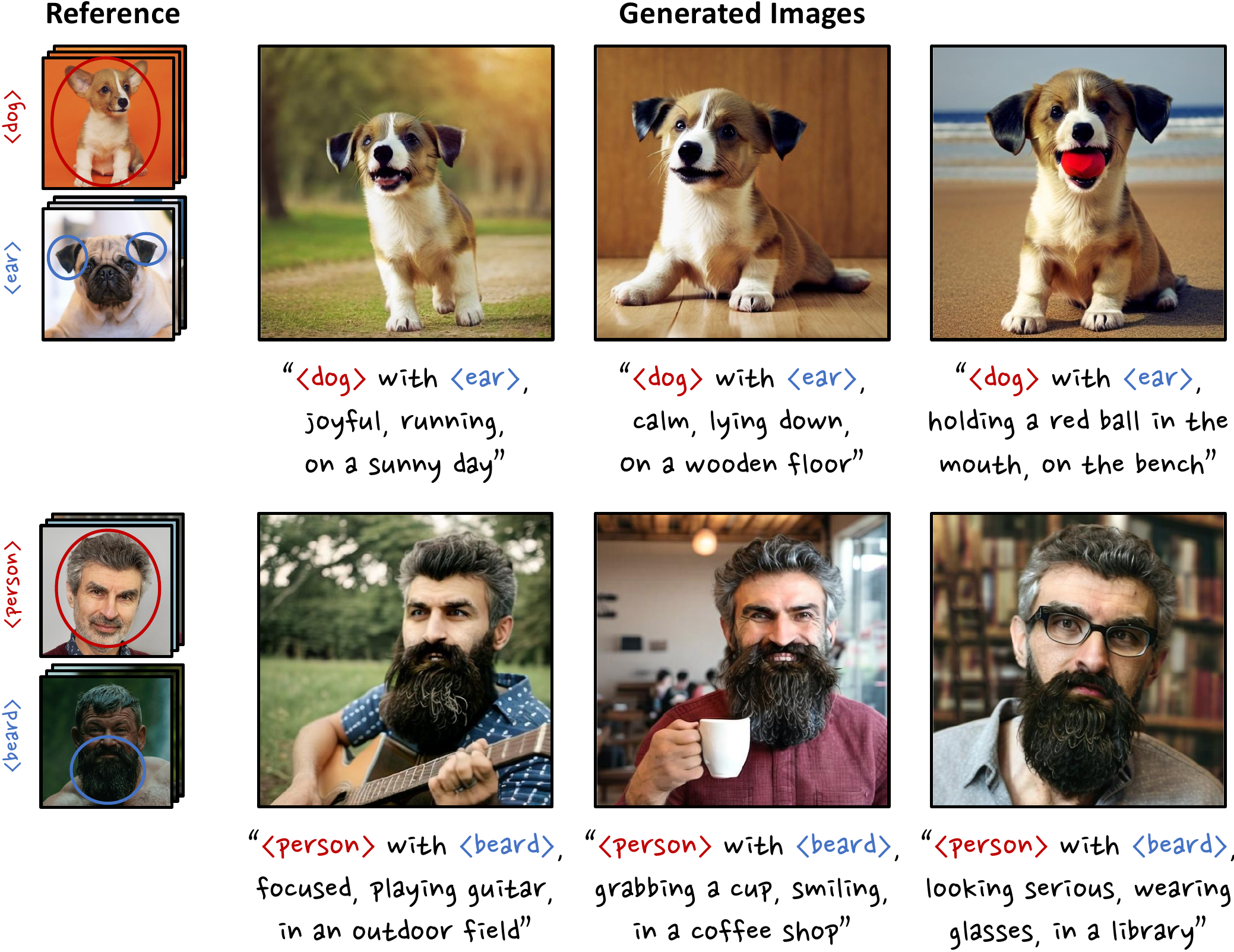}
    
    \vspace{-3mm}
    
    \caption
    {
        \textbf{Generalizability for complex prompts.} 
        We present qualitative results generated with complex text prompts.
        In addition to those well-categorized text prompts, our \method can also follow more complex ones to generate text-aligned images.
    }
    
    \vspace{-3mm}
    
    \label{fig:complex_text}

\end{figure}
\begin{figure}[t]

    \centering
    
    \includegraphics[width=1\linewidth]{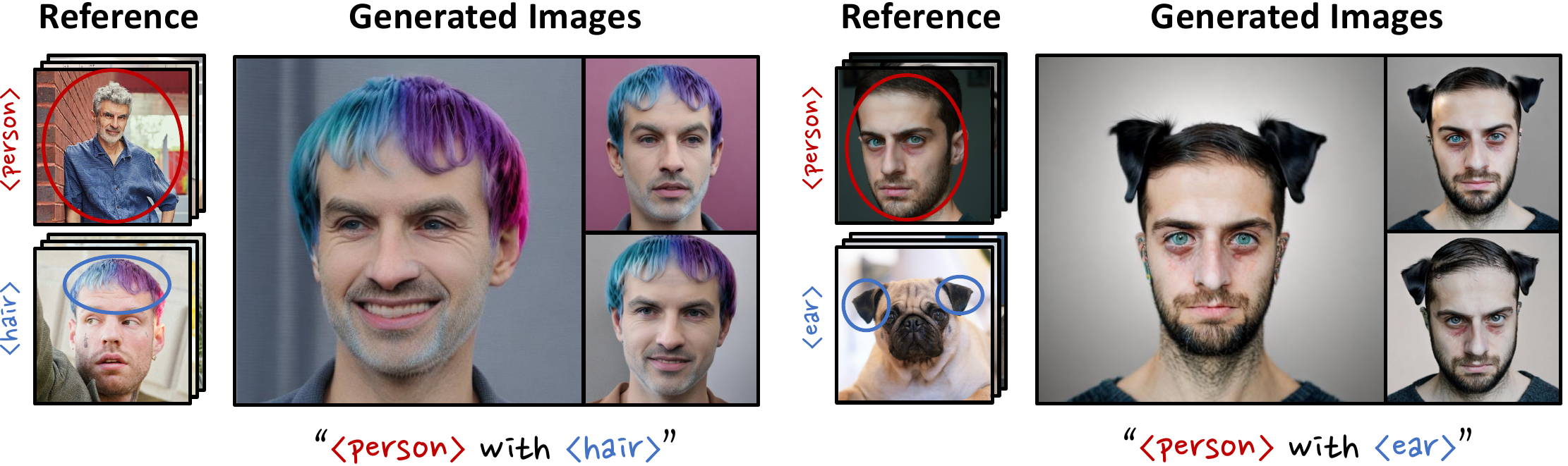}
    
    \vspace{-2mm}
    
    \caption
    {
        \textbf{Generalizability for difficult pairs.} 
        We show the results of two hard cases involving large geometric discrepancy and cross-domain interactions, showing that \method can effectively handle such challenging scenarios.
    }
    
    \vspace{-3mm}
    
    \label{fig:diff_pair}

\end{figure}

\paragraph{Compatibility with Different Backbones.}
\method can also collaborate with other T2I diffusion models as it is a model-independent approach.
In Fig.~\ref{fig:diff_backbone}, we employ \method in other backbones like SD 1.5 \citep{rombach2022high} and SDXL \citep{podell2023sdxl}, showcasing \method can also achieve remarkable results.
Notably, we directly use the original hyperparameter values without further selections, showing the generalizability of \method.

\paragraph{Robustness on Loss Weights.}
In Fig.~\ref{fig:loss_weights}, we analyze the sensitivity of loss weights in Eq.~\ref{eq:total_loss} (\ie, $\lambda_\text{pres}$ and $\lambda_\text{attn}$), since loss weights are often critical for model training.
As we can see, when $\lambda_\text{pres}$ and $\lambda_\text{attn}$ vary within a reasonable range, our \method can consistently attain SOTA performance, revealing its robustness on these hyperparameters.

\paragraph{Performance on Different Numbers of Reference Images.}
In Fig.~\ref{fig:less_ref}, we reduce the number of reference images to analyze the performance variation.
With fewer reference images, \method can still show satisfactory results.
While more reference images could lead to better generalization ability, one reference image per concept/component is enough to obtain a decent result with our \method.

\paragraph{Generalizability to Complex Prompts.}
In comparisons, we have used well-categorized text prompts for systemic evaluation.
Here we further evaluate \method's performance on other complex text prompts involving more complicated contexts.
As shown in Fig.~\ref{fig:complex_text}, \method effectively generates text-aligned images when performing fidelity personalization, showing its ability to handle diverse user needs.

\paragraph{Generalizability to Difficult Pairs.}
We further evaluate \method's performance on challenging pairs, focusing on two cases:
1) large geometric discrepancy, such as ``\textless person\textgreater" in an upper body portrait and ``\textless hair\textgreater" in a profile photo, and 
2) cross-domain interactions, such as ``\textless person\textgreater" and ``\textless ear\textgreater" of dogs.
As shown in Fig.~\ref{fig:diff_pair},  even facing these hard cases, \method can still effectively personalize target concepts and components with high fidelity.

\subsection{Further Applications}
\label{sec:app}

\paragraph{Decoupled Generation.}
After learning from a concept-component pair, \method can also enable decoupled generation.
As shown in Fig.~\ref{fig:applications}(a), \method can generate the target concept and component separately in various and even cross-domain contexts. 
This should be credited to its remarkable ability to capture different-level visual semantics.
Such an ability extends the flexibility of the possible combination between the concept and component. 

\paragraph{Controlling Multiple Components.}
In this paper, we focus on personalizing one concept and one component, because such a setting is enough to cover extensive scenarios, and can be further extended to reconfigure multiple components with an iterative procedure.
However, as shown in Fig.~\ref{fig:applications}(b), our \method also exhibits the potential to control two components simultaneously. 
Handling more components remains a prospective direction of exploring better control over diverse elements for a single concept.

\paragraph{Enhancing Other Generative Tools.}
We demonstrate how \method enhances other generative tools like ControlNet \citep{zhang2023adding}, CSGO \citep{xing2024csgo}, and InstantMesh \citep{xu2024instantmesh} in Fig.~\ref{fig:applications}(c).
\method can integrates seamlessly, furnishing them with an additional ability to control the concept's component in their pipelines.
For instance, working with \method, InstantMesh can conveniently achieve 
fine-grained 3D mesh design, exhibiting the practicability of \method in more creative applications.

\section{Conclusion} \label{sec:conclusion}

We introduce \textit{component-controllable personalization}, enabling precise customization of individual components within concepts. The proposed \textit{\method} uses \textit{Dynamic Masked Degradation (DM-Deg)} to suppress unwanted semantics and \textit{Dual-Stream Balancing (DS-Bal)} to ensure balanced learning. Experiments show that \method sets a new standard in this task, with promising creative applications.
In the future, we would like to extend our approach to broader image and video generation, enabling finer control over multi-level visual semantics for creative generation capabilities.
\section*{Acknowledgments}

We would like to thank Pengzhi Li, Tian Ye, Jinyu Lin, and Jialin Gao for their valuable discussion and suggestions.
This study was supported by the InnoHK initiative of the Innovation and Technology Commission of the Hong Kong Special Administrative Region Government via the Hong Kong Centre for Logistics Robotics.

\bibliographystyle{named}
\bibliography{ijcai25}

\clearpage
\setcounter{page}{1}

\appendix

\section{More Details of Experimental Setup}
\label{app:setup_details}

\subsection{Dataset}
As there is no existing dataset specifically for component-controllable personalization, we curate a dataset from the internet to conduct experiments.
Particularly, unlike previous works \citep{ruiz2023dreambooth, kumari2023multi} that focus on very few categories of concepts, the dataset contains concepts and components from various domains, such as characters, animation, buildings, objects, and animals.
Overall, the dataset consists of 23 concept-component pairs totally with 138 reference images, where each concept/component contains 3 reference images and a corresponding category label.
It is worth noting that the scale of this dataset is aligned with the scale of those datasets used in the compared methods \citep{gal2022image, ruiz2023dreambooth, kumari2023multi, avrahami2023break, safaee2024clic}. 

\subsection{Implementation}
We utilize SD 2.1 \citep{rombach2022high} as the pretrained T2I diffusion model.
As commonly done, the resolution of reference images is set to 512 $\times$ 512.
Besides, the LoRA rank and alpha are set to 32.
To simplify concept learning, we exclude the region of the target component from the segmentation masks of the target concept, \eg, remove the hair from the person in a ``\textless person\textgreater \ + \textless hair\textgreater'' pair.
For the warm-up and DS-Bal stage, we set the learning rate to 1e-4 and 1e-5 and the training steps to 200 and 300.
Moreover, the learning rate is further scaled by the batch size, which is set to completely contain a concept-component pair.
For the cross-attention loss, we follow \citep{avrahami2023break} to average the corresponding cross-attention maps at resolution 16 $\times$ 16 and normalized them to [0, 1].
The model is trained with an AdamW \citep{loshchilov2017decoupled} optimizer and a DDPM \citep{ho2020denoising} sampler.
As done in \citep{avrahami2023break}, the tensor precision is set to float16 to accelerate training.
For a fair comparison, all random seeds are fixed at 0, and all compared methods use the same implementation above except for method-specific configurations.

\subsection{Evaluation}
To generate images for evaluation, we carefully design 20 text prompts covering extensive situations, which are listed in Tab.~\ref{tab:text_prompts}.
These text prompts can be divided into four aspects, including recontextualization, restylization, interaction, and property modification, where each aspect is composed of 5 text prompts.
In recontextualization, we change the contexts to different locations and periods.
In restylization, we transfer concepts into various artistic styles.
In interaction, we explore the spatial interaction with other concepts.
In property modification, we modify the properties of concepts in rendering, views, and materials.
Such a group of diverse text prompts allows us to systemically evaluate the generalization capability of a method.
We generate 32 images per text prompt for each pair, using a DDIM \citep{song2020denoising} sampler with 50 steps and a classifier-free guidance scale of 7.5. 
To ensure fairness, we fix the random seed within the range of [0, 31] across all methods.
This process results in 14,720 images for each method to be evaluated, ensuring a thorough comparison.

\begin{table*}[t]

    \centering
    
     \caption
     {
         \textbf{Text prompts used to generate evaluation images.}
         These text prompts can be divided into four aspects: recontextualization, restylization, interaction, and property modification, covering extensive situations to systemically evaluate the method's generalizability.
         Note that ``\textless placeholder\textgreater" will be replaced by the combination of pseudo-words (\eg, ``\textless tower\textgreater \ with \textless roof\textgreater") when generating evaluation images, and will be replaced by the combination of category labels (\eg, ``tower with roof") when calculating the metric of text alignment.
     }
     
    \setlength{\tabcolsep}{4mm}
    \renewcommand{\arraystretch}{1.3}
    \vspace{-2mm}
    
    \resizebox{0.8\linewidth}{!}
    {
        {
            \begin{tabular}{p{7cm}|p{7cm}}
                \specialrule{0.12em}{0pt}{0pt}
                Recontextualization & Restylization \\
                \hline
                ``\textless placeholder\textgreater, on the beach" & ``\textless placeholder\textgreater, watercolor painting" \\
                ``\textless placeholder\textgreater, in the jungle" & ``\textless placeholder\textgreater, Ukiyo-e painting" \\
                ``\textless placeholder\textgreater, in the snow" & ``\textless placeholder\textgreater, in Pixel Art style" \\
                ``\textless placeholder\textgreater, at night" & ``\textless placeholder\textgreater, in Von Gogh style" \\
                ``\textless placeholder\textgreater, in autumn" & ``\textless placeholder\textgreater, in a comic book" \\
                \specialrule{0.12em}{0pt}{0pt}
            \end{tabular}%
        }
    } 
    
    \vspace{0.3mm}
    
    \resizebox{0.8\linewidth}{!}
    {
        {
            \begin{tabular}{p{7cm}|p{7cm}}
                \specialrule{0.12em}{0pt}{0pt}
                Interaction & Property Modification \\
                \hline
                ``\textless placeholder\textgreater, with clouds in the background" & ``\textless placeholder\textgreater, from 3D rendering" \\
                ``\textless placeholder\textgreater, with flowers in the background" & ``\textless placeholder\textgreater, in a far view" \\
                ``\textless placeholder\textgreater, near the Eiffel Tower" & ``\textless placeholder\textgreater, in a close view" \\
                ``\textless placeholder\textgreater, on top of water" & ``\textless placeholder\textgreater, made of clay" \\
                ``\textless placeholder\textgreater, in front of the Mount Fuji" & ``\textless placeholder\textgreater, made of plastic" \\
                \specialrule{0.12em}{0pt}{0pt}
            \end{tabular}%
        }
    } 
    
    \vspace{-3mm}

    \label{tab:text_prompts}
    
\end{table*}

\subsection{Automatic Metrics}
We utilize four automatic metrics in the aspects of text alignment (CLIP-T \citep{gal2022image}) and identity fidelity (CLIP-I \citep{radford2021learning}, DINO \citep{oquab2023dinov2}, DreamSim \citep{fu2023dreamsim}).
To precisely measure identity fidelity, we improve the traditional measurement approach for personalization. 
This is because a reference image of the target concept/component could contain an undesired component/concept that is not expected to appear in evaluation images.
Specifically, we use Grounded-SAM \citep{ren2024grounded} to segment out the concept and component in each reference and evaluation image. 
Then, we further eliminate the target component from the segmented concept as we have done during training.
Such a process is similar to the one adopted in \citep{avrahami2023break}.
As a result, using the segmented version of evaluation images and reference images, we can accurately calculate the metrics of identity fidelity.

\subsection{User Study}
We further evaluate the methods with a user study.
Specifically, we design a questionnaire to display 20 groups of evaluation images generated by our method and other methods.
Besides, each group also contains the corresponding text prompt and the reference images of the concept and component, where we adopt the same text prompts that are used to calculate CLIP-T.
The results of our method and all the compared methods are presented on the same page.
Clear rules are established for users to evaluate in three aspects, including text alignment, identity fidelity, and generation quality.
Users are requested to select the best result in each group by answering the corresponding questions of these three aspects.
We hide all the method names and randomize the order of methods to ensure fairness. 
Finally, 3,180 valid answers are collected for a sufficient evaluation of human preferences.

\subsection{Compared Methods}
In our experiments, we compare \method with SOTA methods in the domain of personalization, including Textual Inversion (TI) \citep{gal2022image}, DreamBooth-LoRA (DB) \citep{ruiz2023dreambooth}, Custom Diffusion (CD) \citep{kumari2023multi}, Break-A-Scene (BAS) \citep{avrahami2023break}, and CLiC \citep{safaee2024clic}.
We select these methods because TI, DB, and CD are three representatives of personalization frameworks and BAS and CLiC are highly relevant to learning fine-grained elements from reference images.
For TI, DB, and CD, we use the third-party implementation in Diffusers \citep{diffusers}.
For BAS, we use the official implementation.
For CLiC, we reproduce it following the resource paper as the official code is not released.
Unless otherwise specified, method-specific configurations are set up by following their resource papers or Diffusers. 
We empirically adjust the learning rate of CD and CLiC to 1e-4 and 5e-5 respectively, because they perform very poorly with the original learning rates.
For a fair and meaningful comparison, these methods should be adapted to our task setting with minimal modification.
Therefore, for those methods adopting a vanilla diffusion loss, we integrate the masked diffusion loss into them while using the same segmentation masks from \method.

\section{Additional Comparisons}
\label{app:app_add_comp}

\subsection{Detailed Text-Guided Generation}
\label{app:app_add_comp_detailed}
\begin{figure*}[t]

    \centering
    
    \includegraphics[width=1\linewidth]{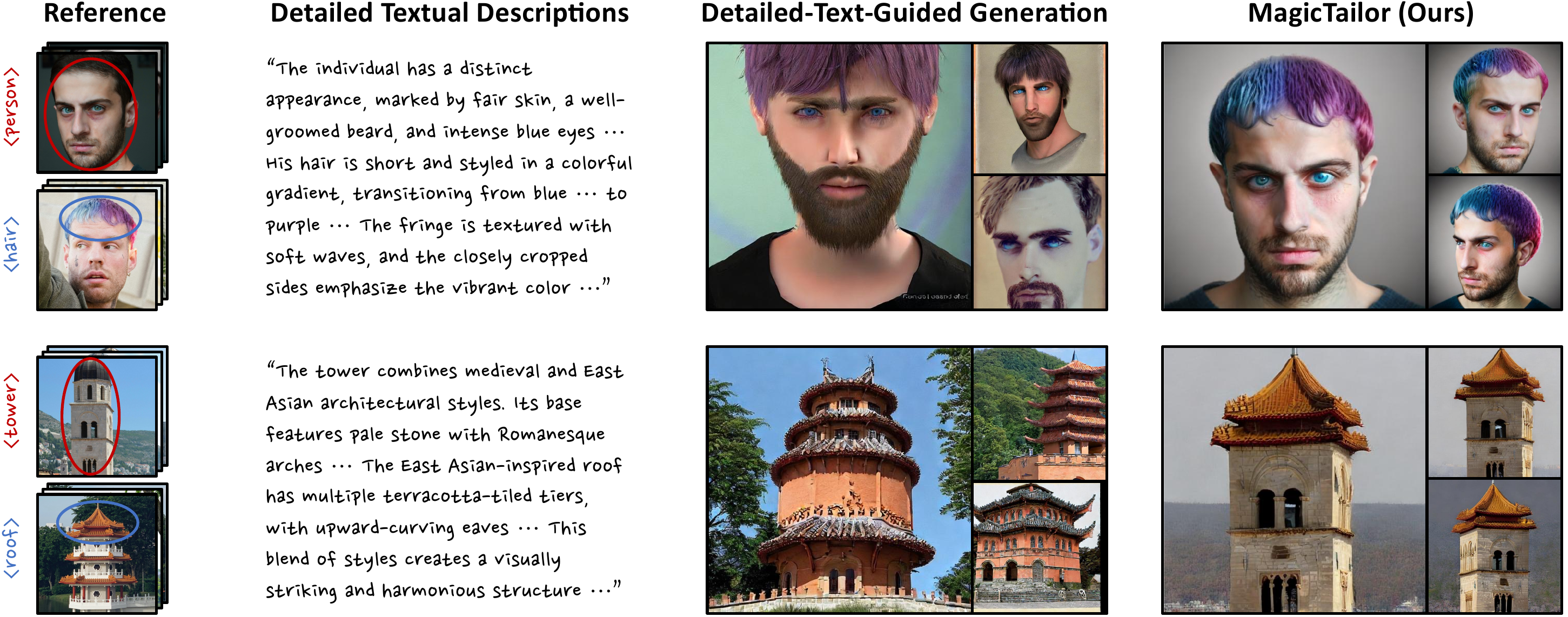}
    \vspace{-6mm}
    \caption
    {
        \textbf{Comparing with detailed-text-guided generation.} 
        We use GPT-4o to generate and merge detailed textual descriptions for the target concept and component, which are fed into Stable Diffusion 2.1 to conduct text-to-image generation. This paradigm cannot perform well and produce inconsistent images, while \method can achieve faithful and consistent generation.
    }
    
    \label{fig:comp_text}
    
    \vspace{-3mm}
     
\end{figure*}

One might wonder if component-controllable personalization can be accomplished by providing detailed textual descriptions to the T2I model.
To investigate this, we separately feed the reference images of the concept and component into GPT-4o \citep{hurst2024gpt} to obtain detailed textual descriptions for them.
The text prompt we used is ``Please detailedly describe the \textless concept/component\textgreater \ of the upload images in a parapraph", where ``\textless concept/component\textgreater" is replaced with the category label of the concept or component.
Then, we ask GPT-4o to merge these textual descriptions using natural language, and input them into the Stable Diffusion 2.1 \citep{rombach2022high} to generate the corresponding images.
Some examples for a qualitative comparison are shown in Fig.~\ref{fig:comp_text}.
As we can see, such an approach cannot achieve satisfactory results, because it is hard to guarantee that visual semantics can be completely expressed by using the combination of text tokens.
In contrast, our \method is able to accurately learn the desired visual semantics of the concept and component from reference images, and thus lead to consistent and excellent generation in this tough task.

\subsection{Commercial Models}
\begin{figure*}[t]

    \centering
    
    \includegraphics[width=1\linewidth]{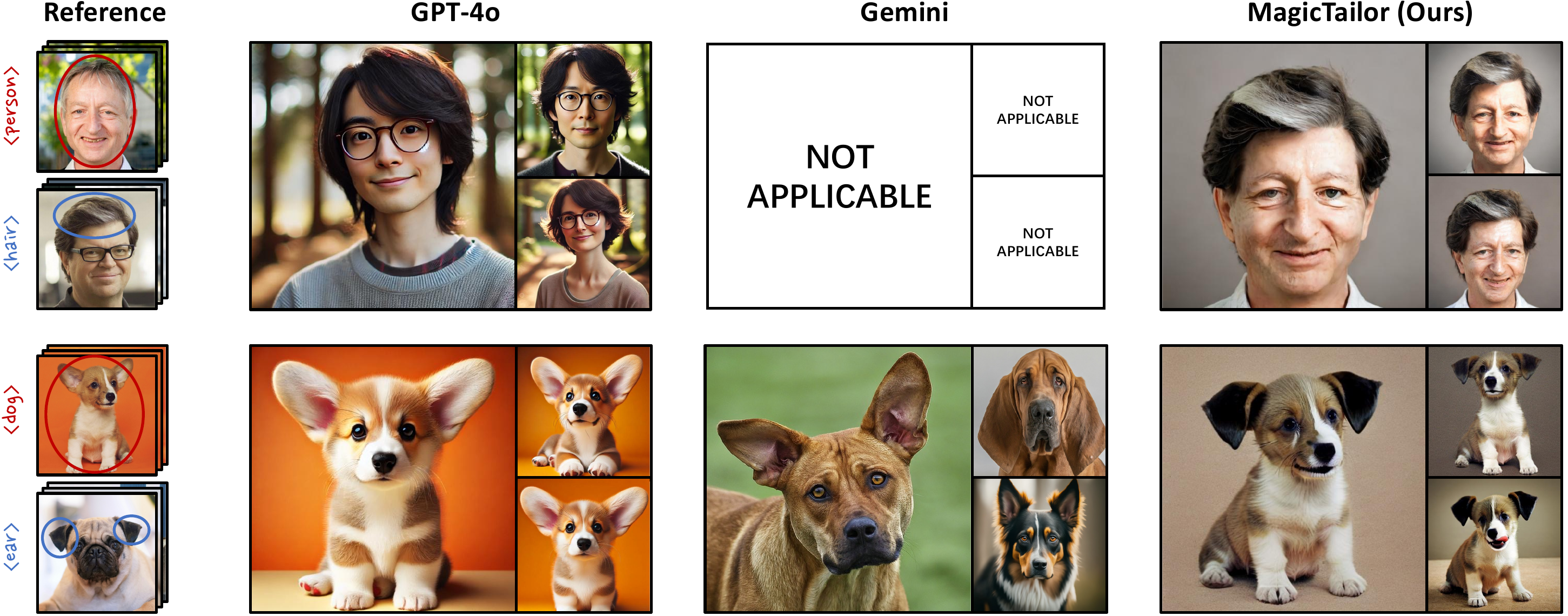}
    \vspace{-6mm}
    \caption
    {
        \textbf{Comparing with commercial models.} 
        We input the reference images of the target concept and component to GPT-4o and Gemini, along with structured text prompts, for conducting image generation.
        Even though capable of handling multiple general tasks, these models still fall short in this task.
        In contrast, our \method performs well using a dedicated framework.
    }
    
    \label{fig:comp_comm}
    
    \vspace{-3mm}
     
\end{figure*}

It is also worth exploring whether existing commercial models, which can understand and somehow generate both text and images by themselves or other integrated tools, are capable of handling component-controllable personalization.
We choose two widely recognized commercial models, GPT-4o \citep{hurst2024gpt} and Gemini 1.5 Flash \citep{team2023gemini}, for a qualitative comparison.
First, we separately feed the reference images of the concept and component into them, along with the text prompt of ``The uploaded images contain a special instance of the \textless concept/component\textgreater, please mark it as \#\textless concept/component\textgreater", where ``\textless concept/component\textgreater" is replaced with the category label of the concept or component.
Then, we instruct them to perform image generation, using the text prompt of ``Please generate images containing \#\textless concept\textgreater \ with \#\textless component\textgreater", where ``\textless concept\textgreater" and ``\textless component\textgreater" are replaced with the category label of the concept and component, respectively.
As shown in Fig.~\ref{fig:comp_comm}, these models struggle to reproduce the given concept, let alone reconfigure the concept's component.
Whereas, our \method achieves superior results in component-controllable personalization, using a dedicated framework designed for this task.
It demonstrates that, even though large commercial models are able to tackle multiple general tasks, there is also plenty of room for the community to explore specialized tasks for real-world applications.

\begin{figure}[t]

    \centering
    
    \includegraphics[width=1\linewidth]{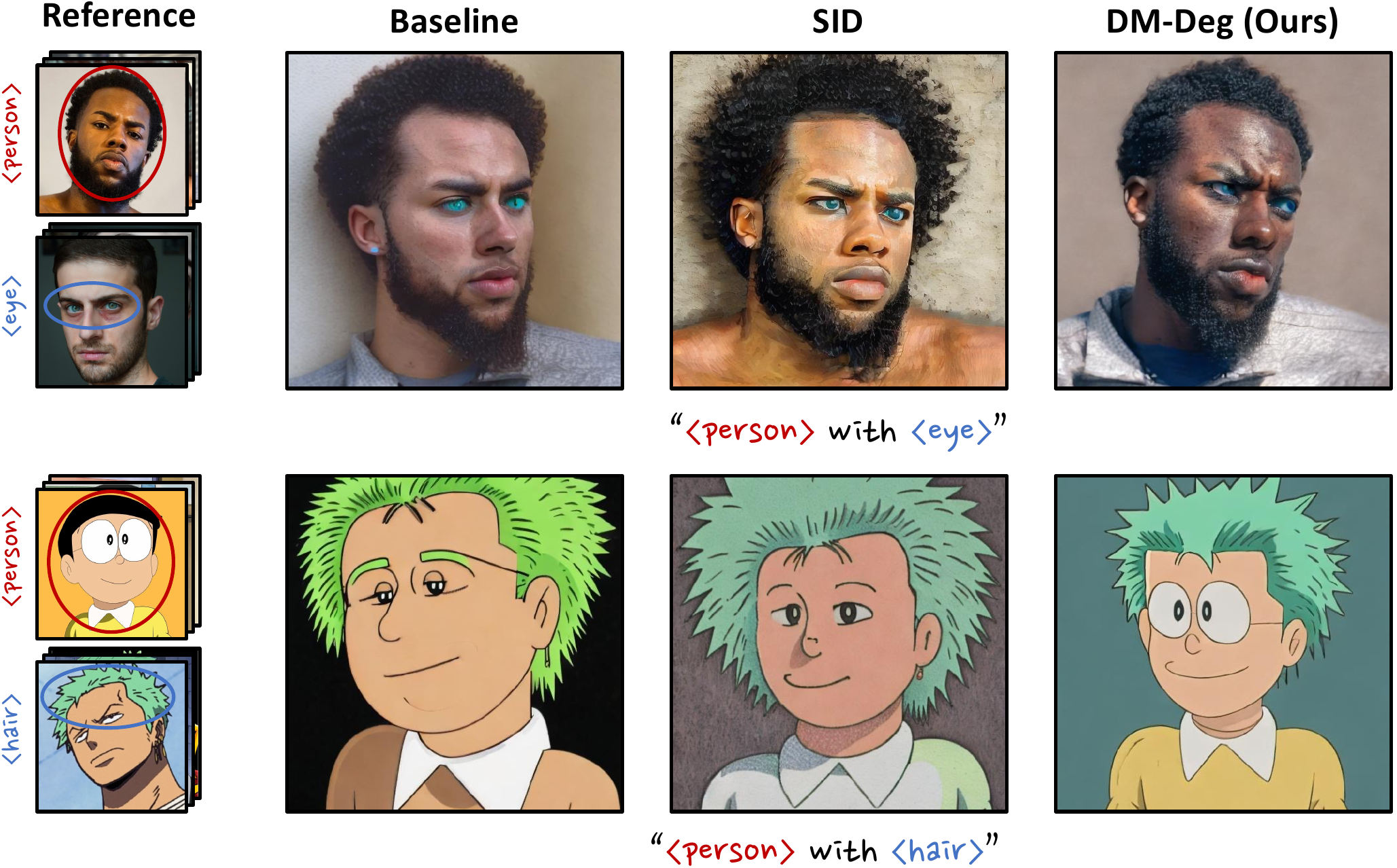}
    
    \vspace{-2mm}
    
    \caption
    {
        \textbf{Ablation of DM-Deg via replacement with SID.} 
        We compare our DM-Deg with SID \citep{kim2024selectively} that aims to produce informative prompts for training.
        Besides, we also present baseline (\ie, removing DM-Deg from \method) results for reference.
        This comparison indicates the effectiveness of our DM-Deg in addressing semantic pollution.
    }
    
    \vspace{-3mm}
    
    \label{fig:sid}

\end{figure}

\section{More Ablation Studies and Analysis}
\label{app:app_add_abl}

\subsection{Dynamic Intensity Matters}
In Tab.~\ref{tab:ablation_deg}, we explore DM-Deg by comparing it with 
1) mask-out strategy;
2) fixed intensity; 
3) linear intensity ($\alpha$ goes from 1 to 0, or from 0 to 1); and
4) dynamic intensity with different $\gamma$.
First, the terrible performance of the mask-out strategy verifies that it is not a valid solution for semantic pollution.
Notably, the descent linear intensity shows better identity fidelity than its ascent counterpart, which aligns with and validates our observations on noise memorization.
Moreover, the dynamic intensity generally shows better results, and it can achieve better overall performance with a proper $\gamma$. 

\subsection{Momentum Denoising U-Net as a Good Regularizer}
In Tab.~\ref{tab:ablation_bal}, we study DS-Bal by modifying the U-Net for regularization as 
1) fixed U-Net with $\beta = 0$ (\ie, the one just after warm-up);
2) fixed U-Net with $\beta = 1$ (\ie, the one from the last step); and
3) momentum U-Net with other $\beta$.
The results show that employing the U-Net with a high momentum rate can yield better regularization to tackle semantic imbalance, thus leading to excellent performance.

\subsection{Necessity of Warm-Up Training}
In \method, we start with a warm-up phase for the T2I model to preliminarily inject the knowledge for the subsequent phase of DS-Bal.
Here we investigate the necessity of such a warm-up phase for generation performance.
In Tab.~\ref{tab:ab_warmup}, when removing the warm-up phase, even though \method could obtain slight improvement in text alignment, it severely suffers from the huge drop in identity fidelity.
This is because such a scheme makes it difficult to construct a decent momentum denoising U-Net for DS-Bal.
Whereas integrated with a warm-up phase, \method can achieve superior overall performance due to the knowledge reserved from warm-up.

\begin{table}[t]
    \centering
         
    \caption
    {
        \textbf{Ablation of DM-Deg.}
        We compare DM-Deg with its variants and the mask-out strategy.
        Our DM-Deg attains superior overall performance on text alignment and identity fidelity.
    }
    \vspace{-2mm} 
    \setlength{\tabcolsep}{2mm}  
    \renewcommand{\arraystretch}{1.3}

    \resizebox{\linewidth}{!}
    {
        \begin{tabular}{lcccc}
            \specialrule{0.12em}{0pt}{0pt}
            Intensity Variants & CLIP-T $\uparrow$ & CLIP-I $\uparrow$ & DINO $\uparrow$ & DreamSim $\downarrow$ \\
            \midrule
            Mask-Out Startegy & 0.270 & 0.818 & 0.760 & 0.375 \\
            \midrule
            Fixed ($\alpha = 0.4$) & 0.270 & 0.849 & 0.800 & 0.297 \\
            Fixed ($\alpha = 0.6$) & 0.271 & 0.845 & 0.794 & 0.310 \\
            Fixed ($\alpha = 0.8$) & 0.271 & 0.846 & 0.796 & 0.305 \\
            Linear (Ascent) & 0.270 & 0.846 & 0.797 & 0.307 \\
            Linear (Descent) & 0.261 & 0.851 & 0.802 & 0.300 \\
            Dynamic ($\gamma = 8$) & 0.266 & 0.850 & 0.806 & 0.289 \\
            Dynamic ($\gamma = 16$) & 0.268 & 0.854 & 0.813 & 0.282 \\
            Dynamic ($\gamma = 64$) & \textbf{0.271} & 0.852 & 0.812 & 0.283 \\
            Dynamic (Ours) & 0.270 & \textbf{0.854} & \textbf{0.813} & \textbf{0.279} \\
            \specialrule{0.12em}{0pt}{0pt}
        \end{tabular}%
    }       

    
    \label{tab:ablation_deg}
    
\end{table}

\subsection{Effectiveness of DM-Deg over Using Informative Training Prompts}
One might be curious about whether it is not necessary to employ the proposed DM-Deg, but perhaps to use informative text prompts during training to provide textual prior knowledge for learning the target concept and component.
To investigate this, we use Selectively Informative Description (SID) \citep{kim2024selectively} with GPT-4o \citep{hurst2024gpt} to construct text prompts for the target concept and component, and then use them for training.
As shown in Fig.~\ref{fig:sid}, such an approach cannot address semantic pollution well, where unwanted visual semantics still affect the personalized concept.
In contrast, DM-Deg effectively prevents semantic pollution by dynamically perturbing those undesired visual semantics, verifying its remarkable significance in this task.

\begin{table}[t]
    \caption
    {
        \textbf{Ablation of DS-Bal.}
        We compare DS-Bal with potential variants, showing its excellence.
    }
    \vspace{-2mm}
    \centering
    \setlength{\tabcolsep}{1.7mm}  
    \renewcommand{\arraystretch}{1.3}   
        
    \resizebox{\linewidth}{!}
    {
    \begin{tabular}{lcccc}
        \specialrule{0.12em}{0pt}{0pt}
        U-Net Variants & CLIP-T $\uparrow$ & CLIP-I $\uparrow$ & DINO $\uparrow$ & DreamSim $\downarrow$ \\
        \midrule
        Fixed ($\beta = 0$) & 0.268 & 0.850 & 0.803 & 0.293 \\
        Fixed ($\beta = 1$) & 0.270 & 0.851 & 0.808 & 0.286 \\
        Momentum ($\beta = 0.5$) & 0.268 & 0.850 & 0.805 & 0.290 \\
        Momentum ($\beta = 0.9$) & 0.269 & 0.850 & 0.808 & 0.288 \\
        Momentum (Ours) & \textbf{0.270} & \textbf{0.854} & \textbf{0.813} & \textbf{0.279} \\
        \specialrule{0.12em}{0pt}{0pt}
    \end{tabular}%
    }
    \vspace{-3mm}        
    \label{tab:ablation_bal}
    
\end{table}
\begin{table}[t]

    \centering

    \caption
    {
        \textbf{Ablations of warm-up.}
        We compare \method with the variant that removes warm-up.
        The results exhibit the significance of the warm-up stage for the framework of \method.
    }
    
    \vspace{-2mm}
    \setlength{\tabcolsep}{2.2mm}  
    \renewcommand{\arraystretch}{1.3}
    
    \resizebox{1\linewidth}{!}
    {
        \begin{tabular}{lcccc}
            \toprule
            Warm-Up Variants & CLIP-T $\uparrow$ & CLIP-I $\uparrow$ & DINO $\uparrow$ & DreamSim $\downarrow$ \\
            \midrule
            w/o Warm-Up & \textbf{0.272} & 0.844 & 0.793 & 0.320 \\
            w/ Warm-Up (Ours) & 0.270 & \textbf{0.854} & \textbf{0.813} & \textbf{0.279} \\
            \bottomrule
        \end{tabular}%
    }    
     
    \vspace{-1mm}
    
    \label{tab:ab_warmup}
    
\end{table}

\subsection{Robustness on Linking Words}
\begin{figure}[t]

    \centering
    
    \includegraphics[width=1\linewidth]{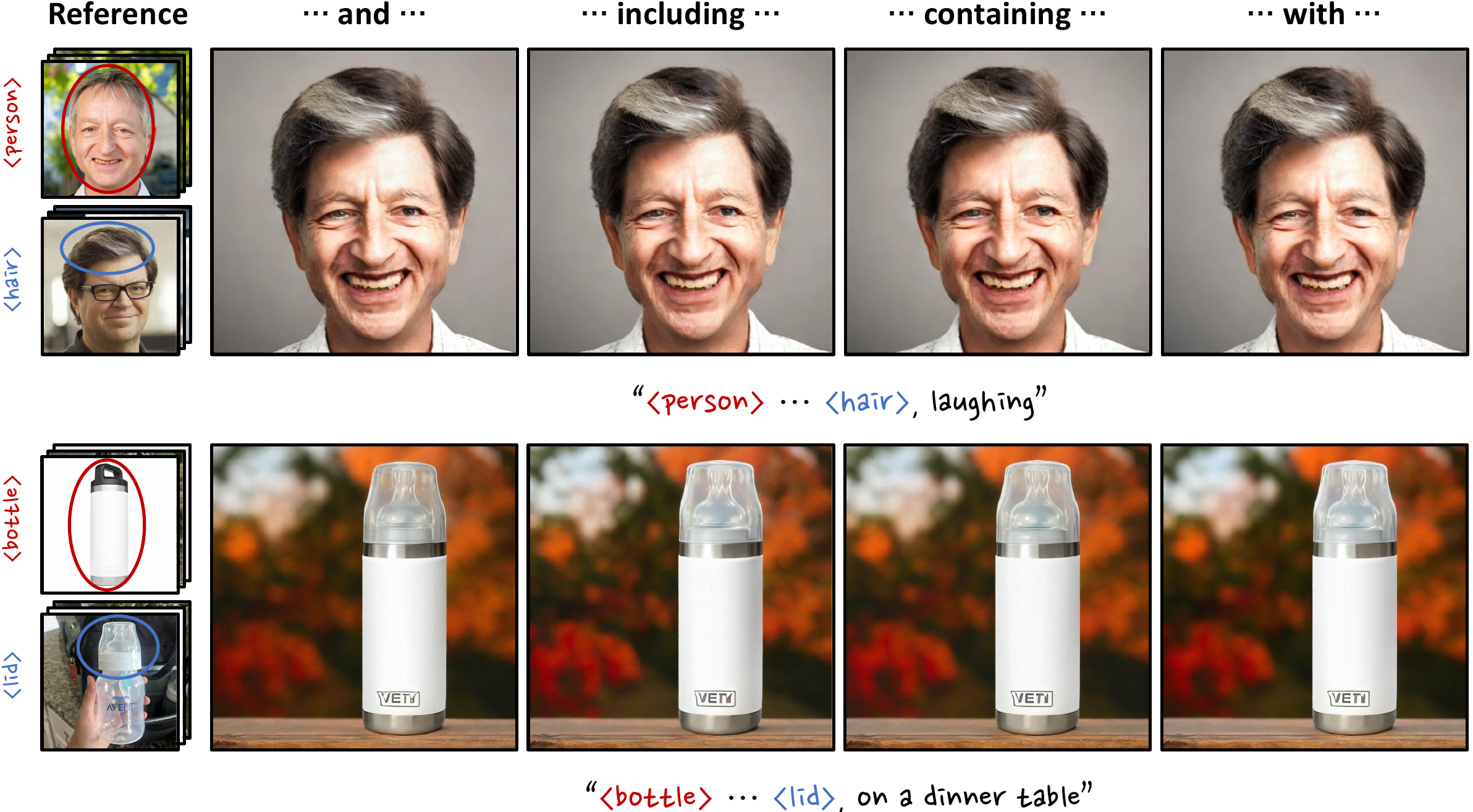}
    
    \vspace{-2mm}
    
    \caption
    {
        \textbf{Ablation of linking words.} 
        We present qualitative results generated with different linking words in text prompts,
        demonstrating the robustness of \method.
    }
    
    \vspace{-3mm}
    
    \label{fig:linking_words}

\end{figure}

Generally, we use ``with" to link the pseudo-words of the concept and component in a text prompt, \eg, ``\textless person\textgreater \ with \textless beard\textgreater, in Von Gogh style''.
Here we evaluate the robustness of our method on different linking words.
We choose several words, which are commonly used to indicate ownership or association, to construct text prompts and then feed them into the same fine-tuned T2I model.
As shown in Fig.~\ref{fig:linking_words}, the generation performance of our \method remains robust regardless of the linking word used, exhibiting its flexibility to textual descriptions.

\section{More Qualitative Results}
\label{app:app_more_qual}
\begin{figure*}[h]

    \centering
    
    \includegraphics[width=1\linewidth]{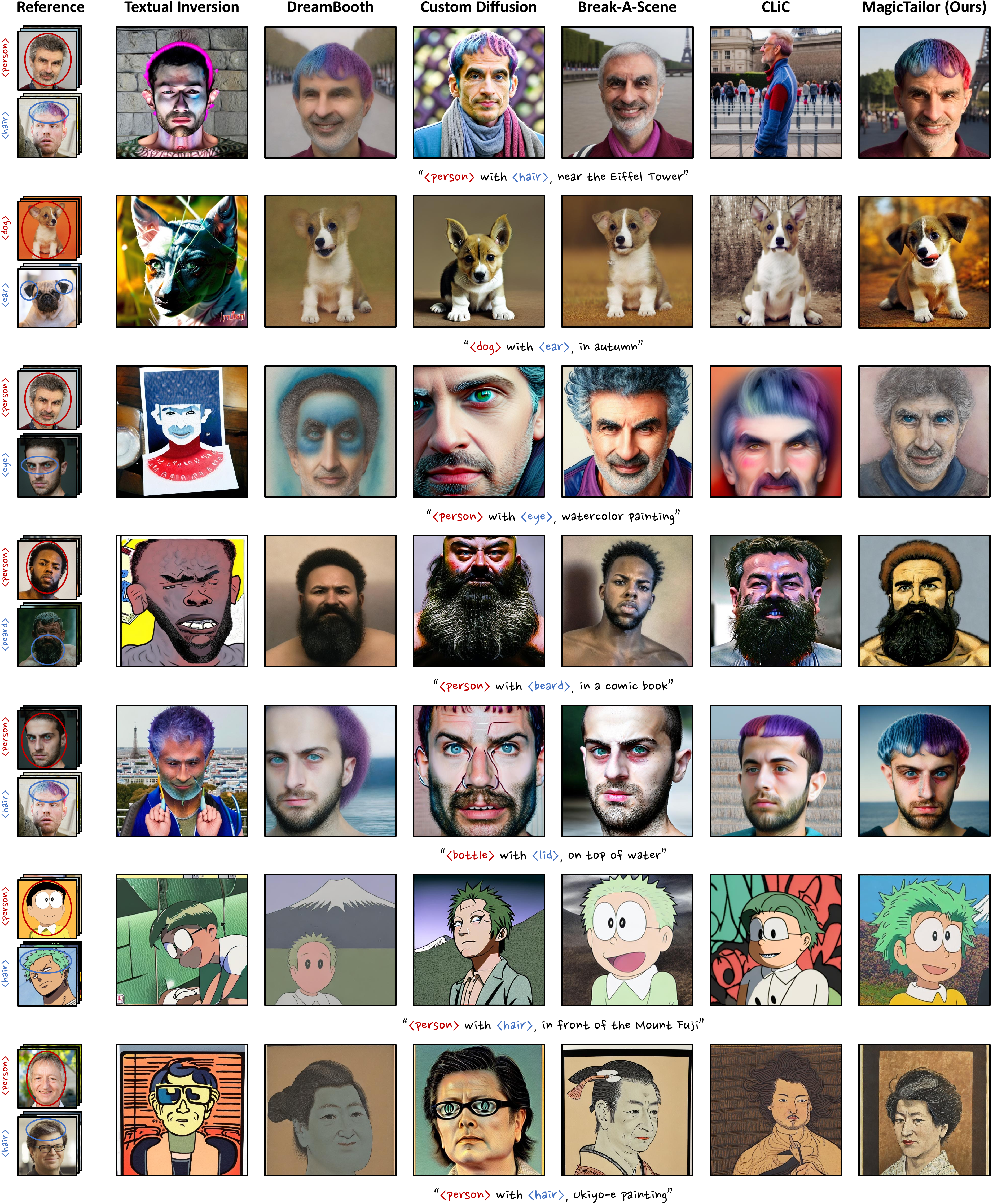}
    \vspace{-6mm}
    \caption
    {
        \textbf{More qualitative comparisons.} 
        We present images generated by our \method and SOTA methods of personalization for various domains including characters, animation, buildings, objects, and animals. 
        \method generally achieves promising text alignment, strong identity fidelity, and high generation quality.
    }
    
    \label{fig:more_qual_results_1}
    
     
\end{figure*}

\begin{figure*}[h]

    \centering
    
    \includegraphics[width=\linewidth]{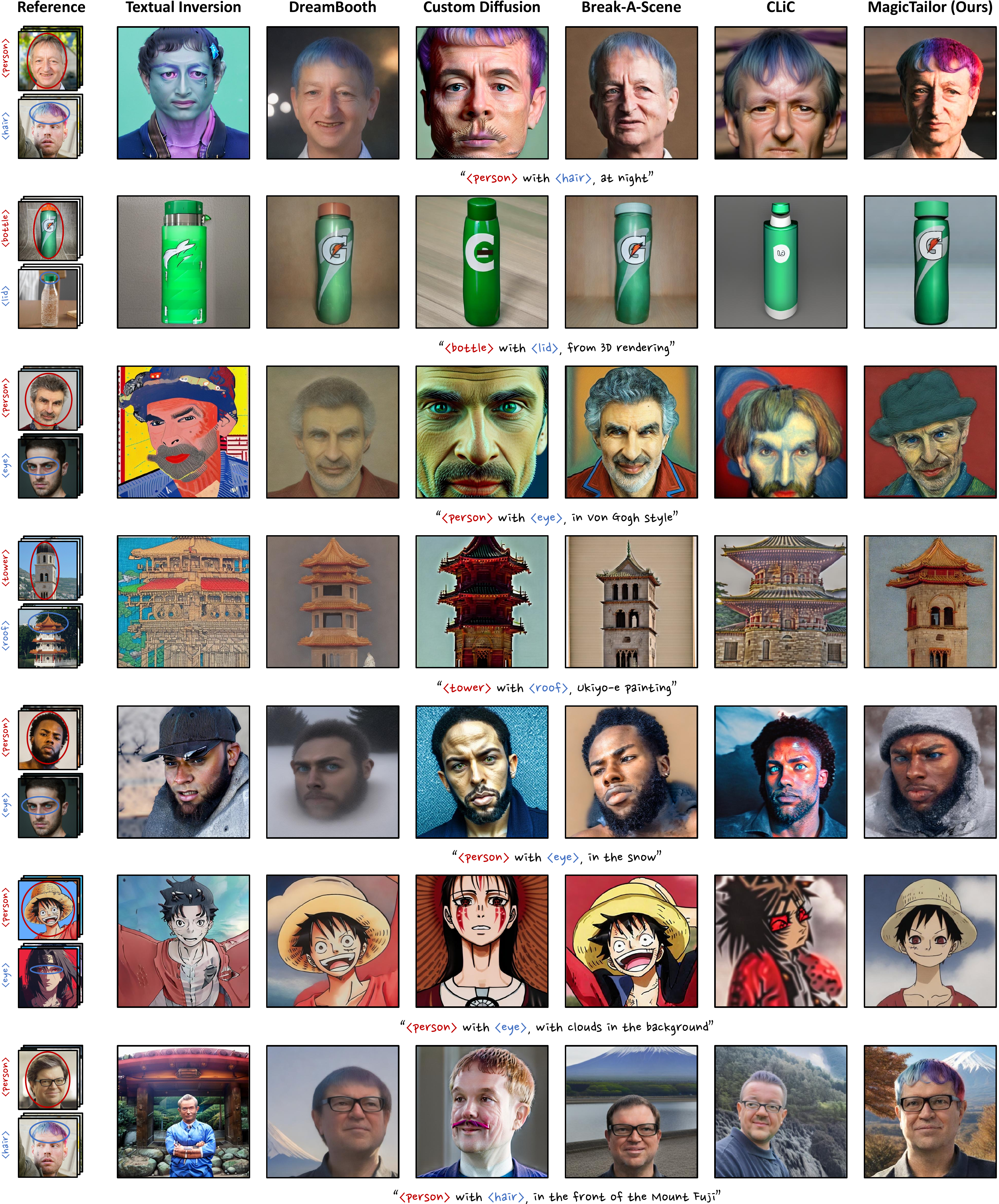}
    \vspace{-6mm}
    \caption
    {
        \textbf{More qualitative comparisons.} 
        We present images generated by our \method and SOTA methods of personalization for various domains including characters, animation, buildings, objects, and animals. 
        \method generally achieves promising text alignment, strong identity fidelity, and high generation quality.
    }
    
    \label{fig:more_qual_results_2}
    
     
\end{figure*}

In Fig.~\ref{fig:more_qual_results_1} \& Fig.~\ref{fig:more_qual_results_2}, we provide more evaluation images for a substantial qualitative comparison.
It can be clearly observed that semantic pollution remains an intractable problem for these compared methods.
This is due to the leak of an effective mechanism to alleviate the T2I model's perception for these semantics.
To address this, our \method utilizes DM-Deg to dynamically perturb undesired visual semantics during the learning phase, and thus achieve better performance. 
On the other hand, the compared methods are also severely influenced by semantic imbalance, resulting in overemphasis or even overfitting on the concept or component.
This is because the inherent imbalance of visual semantics complicates the learning process.
In response to this issue, our \method applies DS-Bal to balance the learning of visual semantics, effectively showcasing its prowess in this tough task.
In summary, the proposed \method effectively addresses both semantic pollution and semantic imbalance through its innovative techniques, DM-Deg and DS-Bal, respectively. These advancements demonstrate its superiority in handling complex visual semantics and achieving remarkable performance in this challenging task.


\end{document}